\PassOptionsToPackage{table,xcdraw}{xcolor}
\documentclass[acmtog]{acmart}

\usepackage{appendix}
\usepackage{soul}
\usepackage{subcaption}
\usepackage{graphicx}

\newcommand{\MYhref}[3][blue]{\href{#2}{\color{#1}{#3}}}%

\AtBeginDocument{%
  \providecommand\BibTeX{{%
    \normalfont B\kern-0.5em{\scshape i\kern-0.25em b}\kern-0.8em\TeX}}}

\setcopyright{acmcopyright}
\copyrightyear{2024}
\acmYear{2024}
\acmDOI{XXXXXXX.XXXXXXX}

\acmPrice{15.00}
\acmISBN{978-1-4503-XXXX-X/18/06}

\acmSubmissionID{367}

\usepackage[normalem]{ulem} 
\usepackage{amsmath} 
\usepackage{tikz}
\usetikzlibrary{decorations.fractals,spy}
\usepackage{multirow}

\usepackage{stfloats}

\begin{document}

\citestyle{acmauthoryear}

\title{A Hierarchical 3D Gaussian Representation for Real-Time Rendering of Very Large Datasets}


\author{Bernhard Kerbl}
\email{kerbl@cg.tuwien.ac.at}
\affiliation{
	\institution{Inria, Université Côte d'Azur}
	\country{France}
}
\authornote{Both authors contributed equally to the paper.}
\affiliation{
	\institution{TU Wien}
	\country{Austria}
}

\author{Andreas Meuleman}
\authornotemark[1]
\email{andreas.meuleman@gmail.com}
\author{Georgios Kopanas}
\email{gkopanas@google.com}
\affiliation{
	\institution{Inria, Université Côte d'Azur}
	\country{France}
}

\author{Michael Wimmer}
\email{wimmer@cg.tuwien.ac.at}
\affiliation{
	\institution{TU Wien}
	\country{Austria}
}

\author{Alexandre Lanvin}
\email{laanvin@gmail.com}
\author{George Drettakis}
\email{George.Drettakis@inria.fr}
\affiliation{
	\institution{Inria, Université Côte d'Azur}
	\country{France}
}


\newcommand{\mparagraph}[1]{%
	\paragraph{{#1}} 
}

\begin{abstract}
Novel view synthesis has seen major advances in recent years, with 3D Gaussian splatting offering an excellent level of visual quality, fast training and real-time rendering.
However, the resources needed for training and rendering inevitably limit the size of the captured scenes that can be represented with good visual quality.
We introduce a hierarchy of 3D Gaussians that preserves visual quality for very large scenes, while offering an efficient Level-of-Detail (LOD) solution for efficient rendering of distant content with effective level selection and smooth transitions between levels. 
We introduce a divide-and-conquer approach that allows us to train very large scenes in independent chunks. We consolidate the chunks into a hierarchy that can be optimized to further improve visual quality of Gaussians merged into intermediate nodes. 
Very large captures typically have sparse coverage of the scene, presenting many challenges to the original 3D Gaussian splatting training method; we adapt and regularize training to account for these issues. We present a complete solution, that enables real-time rendering of very large scenes and can adapt to available resources thanks to our LOD method. We show results for captured scenes with up to tens of thousands of images with a simple and affordable rig, covering trajectories of up to several kilometers and lasting up to one hour.

\end{abstract}

\ccsdesc[500]{Computing methodologies~Rasterization}
\ccsdesc{Computing methodologies~Point-based models}
\ccsdesc{Computing methodologies~Rendering}
\ccsdesc[300]{Computing methodologies~Machine learning approaches}

\keywords{Novel View Synthesis, 3D Gaussian Splatting, Large Scenes, Level-of-Detail}

\begin{teaserfigure}
\includegraphics[width=0.96\textwidth]{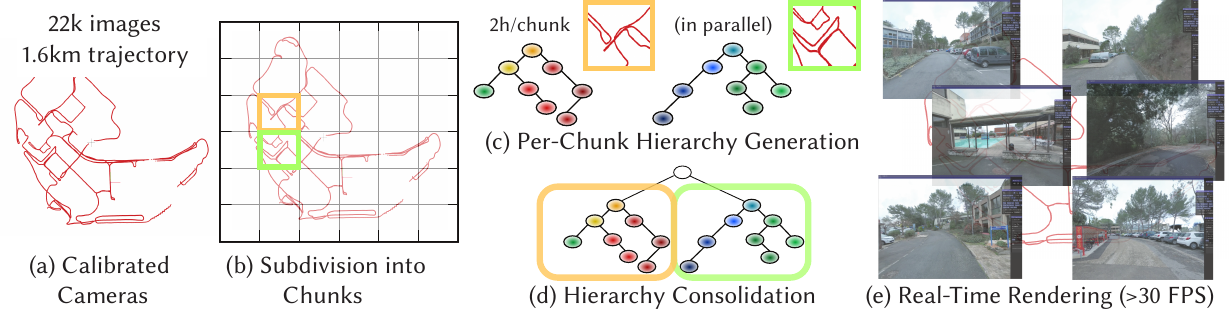}
\caption{
\label{fig:teaser}
(a) Starting from thousands of calibrated cameras, covering a large area, we subdivide the scene into chunks (b). We introduce a 3D Gaussian Splatting hierarchy to allow efficient rendering of massive data, that we further optimize to enhance visual quality (c). We consolidate the hierarchies (d) enabling us to perform real-time rendering of very large datasets. Please see the video for real-time navigation of our large-scale scenes (project page: \MYhref{https://repo-sam.inria.fr/fungraph/hierarchical-3d-gaussians/}{https://repo-sam.inria.fr/fungraph/hierarchical-3d-gaussians/}).
}
\Description[TeaserFigure]{TeaserFigure}
\end{teaserfigure}


\setcopyright{licensedothergov}
\acmJournal{TOG} \acmYear{2024} \acmVolume{43} \acmNumber{4} \acmArticle{62} \acmMonth{7}\acmDOI{10.1145/3658160}

\maketitle

\section{Introduction}

Novel-view synthesis has seen widespread adoption in recent years, in part thanks to the revolution of radiance fields~\cite{ayush}, that provide unprecedented visual quality, and with recent improvements allow interactive or even real-time rendering~\cite{mueller2022instant,reiser2023merf}. 3D Gaussian Splatting (3DGS)~\cite{kerbl3Dgaussians} in particular demonstrates that an explicit primitive-based representation provides an excellent combination of high visual quality, fast training and real-time rendering. 
However, no matter the efficiency of the underlying representation, the available resources for training and rendering limit the size of scenes that can be represented with good quality. We present a new divide-and-conquer solution that allows training and rendering of scenes an order of magnitude larger than most previous methods, by introducing a novel \emph{hierarchy} of 3D Gaussians that allows optimization of interior nodes and provides a level-of-detail renderer.

The vast majority of previous neural radiance field methods cannot scale to very large scenes, due to the implicit nature of the representation, as well as the dependency on grid-like structures used to accelerate computation, that typically have cubic memory growth~\cite{Sun_2022_DVGO}. A few exceptions exist~\cite{meuleman2023localrf}, most notably BlockNeRF~\cite{tancik2022blocknerf}; however, the resources required to train and render such solutions are still extremely high, and real-time rendering has not been demonstrated for these methods. 

On the other hand, 3DGS is a primitive-based rasterization method, opening the possibilty of building on well-understood methodologies for divide-and-conquer and Level-of-Detail (LOD) rendering for large scenes. Unfortunately, the memory requirements of the original 3DGS representation rapidly become too large for even high-end GPUs, making it impossible to render such scenes; training implies an even higher memory overhead. Thus, to train and render very large scenes, we introduce a) a divide-and-conquer method to allow training in smaller pieces, or \emph{chunks}, of the full scene with manageable resources, and preferably allowing parallel processing of chunks, b) a hierarchical structure that will allow fast rendering of distant content, is efficient to build, enables fast cut selection and smooth interpolation between levels, thus providing a good tradeoff between visual quality and speed.

Even though each chunk of a scene could be trivially optimized exactly as in the original 3DGS method, large scene captures, for example from vehicle-mounted rigs, are typically \emph{much sparser} than common radiance field datasets; we thus adapt the optimization of each chunk to this type of input data.
To allow LOD rendering, we present a hierarchy for 3DGS, by defining a merging method for 3DGS primitives based on local geometric and volumetric properties. Our hierarchy allows efficient cut selection, and smooth interpolation between levels. In addition, our hierarchy is built to allow further optimization of intermediate node properties; this second step complements the initial hierarchy construction that is local and purely geometric, and improves overall visual quality. 
Finally, we consolidate the hierarchies of all chunks together, applying a small cleanup step
to remove unecessary hierarchy nodes.
We can then perform real-time rendering using our LOD hierarchy.

We demonstrate our method on several datasets: one provided by Wayve\footnote{http://wayve.ai}, and three we captured ourselves, using a bicycle helmet-mounted rig with 5 or 6 GoPro cameras. 
The datasets cover from 450m up to several kilometers distance, with 5\,800 to 28\,000 images; our method allows real-time navigation in 3D.

Our contributions can be summarized as follows:
\begin{itemize}
  \item A new hierarchy for 3DGS, that allows efficient level selection and interpolation.
  \item A method to optimize the interior nodes of our hierarchy, improving visual quality.
  \item Chunk-based divide-and-conquer training and rendering for large scenes.
\end{itemize}
Our method enables parallel training of chunks of very large scenes and is the first solution with full dynamic LOD, allowing real-time rendering of radiance fields for scenes of such size.
Our solution adapts to available resources and can be used with cheap, consumer-level equipment for capture; this makes capturing and rendering neighborhood-scale scenes accessible to anyone.

We will release our source code, including all supporting code for capture and calibration. We also plan to release our large captures.

\section{Related Work}
Our focus are large-scale scenes, with tens of thousands of input images over several kilometers distance at ground level. 
Despite the impressive progress in novel view synthesis (NVS) and neural rendering in general, very few methods handle environments of the size and complexity we target.
We briefly cover the most relevant work, and focus on results that try to handle scenes with large extent, using meshes, neural fields or point-based representations.

\mparagraph{Image-Based and Mesh-based reconstructions.}
Image-based and mesh-based reconstruction has been used traditionally for scenes of different scales, from small~\cite{chaurasia2013depth} to medium-sized scenes of a room or a few buildings~\cite{buehler2023unstructured,Nishant2023SVS2,Riegler2021SVS,hedman2018deep} up to city-scale data~\cite{bodis2016efficient}. Such methods suffer from a shared drawback; they rely heavily on accurate meshes. Unfortunately, such meshes -- typically obtained by variants of Structure-from-Motion (SfM)~\cite{snavely2006photo} followed by multi-view stereo~\cite{seitz2006comparison} -- are known to fail in challenging cases of vegetation, thin structures and non-Lambertian or texture-less materials. All these appear frequently in a standard cityscape when captured from street level.
Other approaches achieve fast rendering~\cite{liu2023neuras, Riegler2021SVS} with features on a 3D mesh scaffold built via SfM. The features can be optimized when rendering~\cite{liu2023neuras} or extracted from images before an on-surface aggregation step \cite{Riegler2021SVS}. Due to their heavy reliance on meshes, they tend to recover fine structures less accurately and, like other methods that consider the entire data simultaneously, they are not arbitrarily scalable. 
There has been extensive work on capturing urban data~\cite{UrbanScene3D,zhang2021continuous,zhou2020offsite} but the focus tends to be on aerial capture rather than the street level data we consider here.

\mparagraph{Radiance Field Reconstruction and Rendering.}
Neural Radiance Fields (NeRFs)~\cite{mildenhall2020nerf} recover a volumetric radiance field for bounded scenes, usually centered around a single object. Mip-NeRF~\cite{barron2021mip} allows for proper anti-aliasing to handle multi-scale observations. In parallel, NeRF++~\cite{kaizhang2020nerfpp} lifted the constraints of a bounded scene and Mip-NeRF 360~\cite{barron2022mipnerf360} applied the benefits of a properly anti-aliased representation in unbounded object-centric scenes. These methods reconstruct a small area of interest with high quality while compressing the background using space contraction. Voxel-based representations~\cite{Sun_2022_DVGO, karnewar2022relu} have been extensively studied to improve the optimization and rendering speed of these methods. But the field-based, implicit nature of all these methods naturally suggest a tradeoff between the quality of the reconstruction and the cubic growth of the voxel representations of scenes. This tradeoff can be partly addressed through compression and empty space skipping using, e.g., hash-grids~\cite{mueller2022instant} or tensor decomposition~\cite{tensorf}. F2-NeRF~\cite{wang2023f2nerf} goes further, lifting the assumption of an object-centric scene, and warps space to allocate the capacity of the representation efficiently, depending on an arbitrary camera trajectory. Other NeRF-based methods~\cite{barron2023zipnerf, duckworth2023smerf, wu2022snisr, zhang2022nerfusion} show the ability to scale in indoor scenarios up to apartment level; our city-scale datasets are an order of magnitude larger both in extent and number of images.

Recently-introduced 3D Gaussian Splatting~\cite{kerbl3Dgaussians} is the first method to achieve high visual quality in unbounded scenes while maintaining fast training and real-time rendering. For our purposes, the important benefit of 3DGS is that it forgoes implicit field-based solutions and uses a primitive based representation~\cite{keselman2022fuzzy} that does not need to pre-allocate data structures before optimization. This allows arbitrary camera paths and dynamic allocation of representational capacity where necessary. While arbitrary camera paths do not pose a problem for 3DGS, arbitrarily large scenes will eventually saturate resources, making the use of 3DGS infeasible in such scenes. We address the resource saturation problem by introducing the first 3DGS hierarchy that can be optimized and provides an efficient Level-of-Detail solution together with a divide-and-conquer technique to subdivide the environment.

\mparagraph{Level-of-Detail Rendering.} Level-of-detail approaches are a well-established part of computer graphics~\cite{luebke2003level}. More recently, similar ideas have been applied in the context of NeRFs~\cite{xiangli2022bungeenerf, takikawa2022variable} and learned Signed Distance Functions~\cite{takikawa2021neural}. 
In real-time rendering applications, LODs can regulate the amount of detail that is displayed at any point in time, based on heuristics or targeted resource budgets.
LODs provide essential optimization and often form the building blocks for truly scalable rendering solutions \cite{karis2021nanite}. 
The ability to adjust the amount of detail enables flexibility for developers and users alike.
LODs have been proposed for point-based representations~\cite{qsplat,seqtrees}, but the dual volumetric/rasterizable nature and the optimization processs of 3DGS poses specific challenges that we address here.
While even simple approximations of scenes (e.g., use of voxel grids or quantization \cite{https://doi.org/10.1111/cgf.14345}) can already qualify as an LOD, a complete approach suitable for interactive scenarios must address three challenges simultaneously: the generation of an LOD structure with multiple levels of detail, a policy to select the appropriate level for a given view, and the ability to transition between them without causing disruptive artifacts.
Our hierarchy fulfills these requirements and provides a solution fit for real-time rendering.

\mparagraph{City Scale Reconstruction.}
Few research solutions can handle city-scale scenes at ground level, because of the complexity of the problem.
In addition, abundant high-quality data is not publicly available. We hope the planned public release\footnote{Pending Data Protection Officer approval.} of our datasets will help on this front.  
BungeeNeRF~\cite{xiangli2022bungeenerf} requires specific data during the progressive optimization: from satellite captures to closer views. It is therefore inappropriate for ground-level city scale data captures (even if it sometimes includes additional aerial footage). NeuRas~\cite{liu2023real} uses an MVS reconstruction and an optimized neural texture to model urban scenes that span a few seconds of driving footage, which is significantly smaller than the goal of this paper. DrivingGaussians~\cite{zhou2023drivinggaussian} builds on top of 3DGS, but they reconstruct scenes from nuScenes~\cite{caesar2022nuplan} and KITTI360~\cite{Liao2022PAMI} at extents that easily fit in their hardware configuration.

Divide-and-conquer solutions partition the scene in independent blocks or chunks; we also adopt this strategy. In the context of radiance fields, KiloNeRF~\cite{reiser2021kilonerf} introduced such a solution to address performance issues, but later the same idea was employed for large-scale scenes~\cite{tancik2022blocknerf, Turki_2022_meganerf, meuleman2023localrf, dhiman2023stratanerf}. The method closest to ours is BlockNerf~\cite{tancik2022blocknerf} that partitions the scene into parts that overlap by 50\%, and computes a NeRF for each part. However, the NeRF method demonstrated is very slow to train and render.
In contrast, we present a method that allows fast training per chunk after a short coarse initialization, and most importantly allows real-time rendering.

\section{Overview and Background}
\label{sec:defs}

We address the challenge of optimizing very large scenes to create a hierarchical radiance field that can be rendered in real-time.
We first need to address the problem of limited available resources for optimization, given data that is typically much sparser than traditional radiance field captures presented to date. We do this by subdividing the scene (Fig.~\ref{fig:teaser}(a)) into a set of chunks (b). A first, coarse \emph{scaffold} is created by optimizing a fixed, small number of 3D Gaussians for the entire scene. 
We then optimize each chunk independently (c), introducing several improvements to the 3DGS optimization to handle sparse data (Sec. \ref{sec:chunk-train}). 
To render distant content efficiently, we introduce a new 
3DGS hierarchy (d) for each chunk, as well as an optimization method for the properties of interior nodes that improves visual quality overall.
The optimized hierarchies for each chunk are compressed, and consolidated into a complete hierarchical representation of the entire scene (e). 
The hierarchy can then be used for LOD-based real-time rendering.

After a brief background overview of 3DGS, we introduce our hierarchy (Sec.~\ref{sec:methodlod}), then describe how we optimize the interior nodes (Sec.~\ref{sec:compact-hier}) and finally we discuss the chunk-based optimization for large scene training (Sec.~\ref{sec:methodlarge}). 

\subsection{Background}

3DGS creates a scene representation based on volumetric primitives, that each have the following set of parameters: position (also referred to as mean) $\mu$, covariance matrix $\mathbf{\Sigma}$ that, in practice, is decomposed into scale and rotation, opacity $\mathrm{o}$, and spherical harmonics (SH) coefficients to represent appearance, or view-dependent color.

The 3D primitives are projected to 2D screen space, and rasterized using $\alpha$-blending. The $\alpha$-blending weights are given as
\begin{equation}
	\label{eq:alpha}
\alpha ~=~ \mathrm{o} \mathrm{G}
\end{equation}
with the projected Gaussian contribution on pixel $(x,y)$ given as:
\begin{equation}
	G(x,y)~=~e^{-\frac{1}{2}([x, y]^T-\mu')^T \mathbf{\Sigma}'^{-1}([x, y]^T-\mu')}
\end{equation}
where $\mu'$ and $\mathbf{\Sigma}'$ are the projected 2D mean and covariance matrix.
The combined effect of converting SHs to per-view color values and $\alpha$-blending them recreates the appearance of the captured scene.

\section{Hierarchical LOD for 3D Gaussian Splatting}
\label{sec:methodlod}

Level-of-detail (LOD) solutions are critical when handling large scenes to allow efficient rendering of massive content;
Our goal is thus to create a hierarchy that represents the primitives generated by the original 3DGS optimization. 
Following traditional LOD methods in graphics, we need to 1) 
find candidate 3DGS primitives and 
define how to merge them
into intermediate nodes, 2) provide an efficient way to determine a cut in the hierarchy that provides a good tradeoff between quality and speed, and 3) a smooth transition strategy between hierarchy levels.

\subsection{Hierarchy Generation}

We create a tree-based hierarchy with interior and leaf nodes for each chunk. Every node is associated with a 3D Gaussian, which is either a leaf node coming from the original optimization, or a merged interior node.
Our requirements for intermediate nodes are that they should: 1) Maintain the same fast rasterization routine as leaf nodes; 2) Represent the appearance of children as accurately as possible. We thus need to define intermediate nodes that are 3D Gaussians
that have all the attributes of 3DGS primitives, i.e., mean $\mu$ (position), covariance $\mathbf{\Sigma}$, SH coefficients and opacity (see Sec.~\ref{sec:defs}). 

For mean and covariance, there is exhaustive literature related to Gaussians that we build on to define our merging process. 
Specifically, to merge $N$ Gaussian primitives with means $\mu_i^{(l)}$ and covariances $\mathbf{\Sigma}_i^{(l)}$ of level $l$ s.t.\ the 3D Kullback-Leibler divergence between the merged node and its children's weighted distributions is minimized, we use~\cite{goldberger2004hierarchical,jakob2011progressive}:
\begin{align}
	\label{eq:merge}
	\mu^{(l+1)} &= \sum_i^N w_i \mu_i^{(l)},\\
	\Sigma^{(l+1)} &= \sum_i^N w_i (\Sigma_i^{(l)} + (\mu_i^{(l)} - \mu^{(l+1)})(\mu_i^{(l)} - \mu^{(l+1)})^T) 
\end{align}

\noindent
where $w_i$ are normalized weights, i.e., $w_i =\frac{w'_i}{ \sum_i^N w'_i}$.
We next define the unnormalized merging weights $w'_i$ that are proportional to the contribution each child Gaussian has to the created parent. 

\begin{figure}
	\includegraphics[width=\columnwidth]{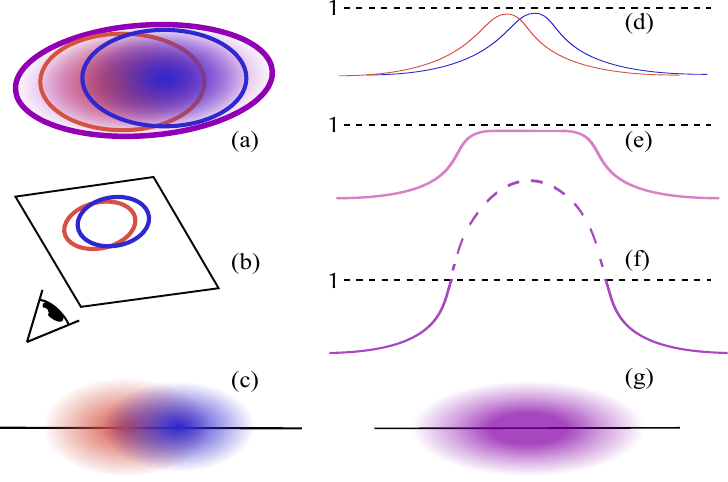}
	\caption{
		\label{fig:hierarchy}
		(a) The blue and red 3D Gaussians are leaf primitives that are projected to 2D (b). We visualize a scanline (black line in (c)) in 2D and plot the corresponding $\alpha$-blending weights $\alpha_r$ and $\alpha_b$ for red and blue respectively. The cumulative effect of blending according to their opacity (Eq.~\ref{eq:blend}) is shown in (e); we see that the effect is a non-Gaussian cumulative fall-off. We want to create an intermediate node to represent the two leaves, shown in purple (a). Taking a scanline through the purple projected intermediate node (g), we show that the \emph{falloff} value we introduce to replace \emph{opacity} achieves a similar slower fall-off effect (f); however the value can be larger than 1, which we clamp appropriately during $\alpha$-blending (see text).
	}
\end{figure}

To find these weights, we reason in screen space for projected 2D Gaussians. 
For an isolated Gaussian primitive $g_i$ 
with color $\mathrm{c_i}$ and opacity $\mathrm{o_i}$ 
the contribution $C_i(x,y)$ to an image position $(x,y)$ is:
\begin{equation}
\label{eq:blend}
	C_i(x,y) ~=~\mathrm{o_i}~\mathrm{c_i} G(x,y)
\end{equation}
The contribution $C_i$ to the entire image is then:
\begin{align}
	C_i ~&=~ \mathrm{o_i}~\mathrm{c_i} \int_X \int_Y G(x,y) \\
	 ~&=~ \mathrm{o_i}~\mathrm{c_i} \sqrt{(2\pi)^2 |\mathbf{\Sigma}'|}
\end{align}
\noindent
from the properties of Gaussians.
To derive our weights, we make some simplifying assumptions: Gaussians are nearly isotropic with little overlap and low perspective distortion. In this simplified case for two Gaussians, we want to ensure that the contribution of the parent Gaussian $g_p$ is equal to that of the two children $g_1,g_2$. We thus need $C_p$ of the parent to be equal to the combined contribution of the two children $C_1 ~+~ C_2$. If we solve for the weights required, we obtain the following expression for $w_i$ that can be used in Eq.~\ref{eq:merge}, ignoring constant factors and color that are not relevant for the weights:
\begin{equation}
	w'_i ~=~ \mathrm{o_i} \sqrt{|\mathbf{\Sigma}'_i|} 
\end{equation}
In practice, since the square root of the determinant of a Gaussian's 2D covariance is proportional to the (projected) surface of the corresponding 3D ellipsoid, we compute the surface $S_i$ of each ellipsoid instead of $\sqrt{|\mathbf{\Sigma}'_i|}$.

We can similarly use the weighted average of the SH coefficients for the merged node using the same weights:
\begin{equation}
	SH^{(l+1)} = SH_1^{(l)} w_1 + SH_2^{(l)} w_2.
\end{equation}

\noindent
We can also use the weights to merge opacity. However, our merging strategy changes the semantics of the opacity property for intermediate nodes.
Consider the red and blue 3D Gaussians shown in Fig.~\ref{fig:hierarchy}(a), that are rasterized onto the screen (b). 
The result of multiple blended close-by Gaussians can result in slower-than-Gaussian falloff; see for example
the plot of $\alpha$-blended opacity (c) with corresponding blending weights $\alpha_r$ and $\alpha_b$ across a scanline in screenspace (d). We see that each individual primitive has a standard Gaussian fall-off. If we plot the cumulative effect of $\alpha$-blended opacity, we see
a slower fall-off effect (e). 
Using the weights $w_i'$, we can
plot a merged Gaussian $p$ with opacity $\frac{\sum_i^Nw_i'}{S_p}$ (f) and resulting contribution along the scanline (g); we see that the slow fall-off effect is maintained. However, this quantity can now be larger than 1: instead of opacity, we thus call this value \emph{falloff} for intermediate nodes. It behaves as opacity during rendering, but resulting $\alpha$ values are clamped to 1.

Gaussians in a typical scene may violate the above assumption of isotropy. In Sec. \ref{sec:post-optim}, we describe additional measures to address these cases and improve quality of higher LOD levels in such cases.

We now have a merging procedure for our 3D Gaussians. 
Given a set of 3DGS primitives, we first build an axis-aligned bounding box (AABB) Bounding Volume Hierarchy (BVH) top-down over them.
We start from an AABB that encloses all Gaussian primitives, using  $3\times$ their stored size to capture their extent (using the cutoff from the original paper~\cite{kerbl3Dgaussians}).
The initial AABB is the root node of the BVH.
To obtain the child nodes, we recursively perform a binary median split on the current node. First we project the means of all Gaussians in a group onto the longest axis of their bounding box. We then partition the group by assigning each primitive based on the position of its projected mean with respect to the median of all projections.
The resulting BVH tree ensures that the children of each interior node will be spatially compact.
We then compute the intermediate node Gaussians from their respective children, starting from the leaves and recursively merging up the tree.

\begin{figure}
	\centering
	\includegraphics[width=0.9\linewidth]{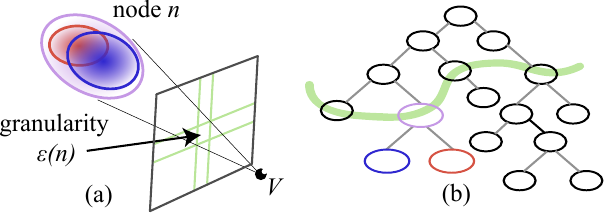}
	\caption{
		\label{fig:gran-cut}
		(a) The granularity $\epsilon(n)$ of the green node $n$ is defined as the projected screen size of the node. (b) Nodes satisfying target granularity $\tau_{\epsilon}$ of e.g., 1 pixel or less are included in the cut for a given view $V$. In practice, $\tau_{\epsilon}$ is a threshold for projected AABB axis lengths $\epsilon(n)$ of each node $n$.
	}
\end{figure}

\subsection{Hierarchy Cut Selection and Level Switching}
\label{sec:select}

Given a hierarchical tree structure for a 3DGS model, and a view $V$ we select a cut through the tree that will maximize rendering performance while maintaining visual quality.
We first define the \emph{granularity} $\epsilon(n)$ of a given hierarchy node $n$ as the projected size on the screen for a given view (Fig.~\ref{fig:gran-cut}(a)).
Specifically, we use the bounding box over leaf Gaussians contained in a node, then take the largest dimension of the bounding box and use this to compute the projected size. 
We find the cut by identifying the nodes whose projected bounding box is smaller than a given \emph{target granularity} $\tau_{\epsilon}$ on screen, e.g., 1 pixel. 
Based on the generation of the hierarchy nodes and its bounds, the AABB of a parent can never appear smaller than one of its children. 
This enables us to find the appropriate cut in linear time or, in a massively parallel setting, in constant time per node: if the bounds of a node $n$ fulfill the granularity condition, but its parent's do not, then node $n$ is chosen for the given setting and included in the cut (green curve in Fig.~\ref{fig:gran-cut}(b)). Note that either all children or the parent are chosen by this logic.

A key element of any hierarchical rendering solution is the ability to allow smooth transitions between hierarchy levels. 
We achieve smooth transitions by interpolation on the individual Gaussian attributes. 
When a node is no longer the best fit for the current target granularity, it is replaced by its children by 
interpolating between parent and children Gaussian attributes.
The cut returned by hierarchy selection with target granularity $\tau_{\epsilon}$ will contain nodes that (over-)fulfill the criterion. 
Interpolation weights are selected by evaluating the granularity $\epsilon(n)$ of each node in the cut, as well as the granularity $\epsilon(p)$, where $p$ is the parent of $n$.
The interpolation weight $t_n$ is then found as:
\begin{equation}
\label{eq:t_interp}
t_n~=~\frac{\tau_{\epsilon} - \epsilon(n)}{\epsilon(p) - \epsilon(n)}.
\end{equation}

\begin{figure}
	\hspace{\fill}
	\begin{subfigure}[b]{0.55\linewidth}
		\includegraphics[width=\linewidth]{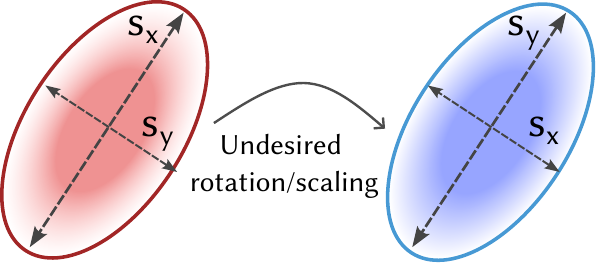}
		\caption{}
	\end{subfigure}
	\hfill
	\begin{subfigure}[b]{0.38\linewidth}
		\includegraphics[width=\linewidth]{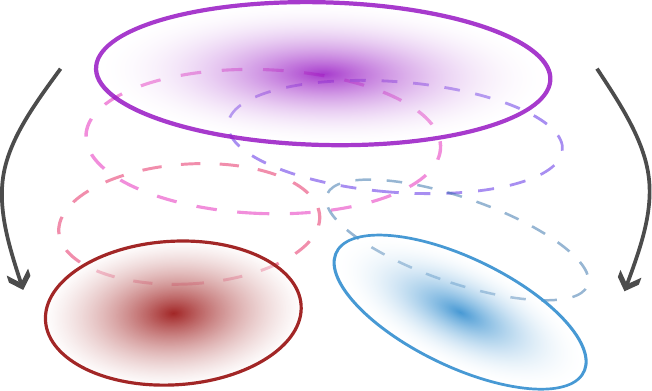}
		\caption{}
	\end{subfigure}
	\hspace{\fill}
	\caption{
		\label{fig:interp}
		(a) The ambiguity of the rotation axes of Gaussians can result in undesired rotations when interpolating between nodes.
		(b) When switching nodes, the two children are rendered with the same parameters as the purple parent, and progressively interpolated towards their separate values.
	}
\end{figure}

Position, color and spherical harmonics can be interpolated linearly using the interpolation weights.
For covariance, we found
that interpolating scale and rotation separately results in superior appearance to covariance matrix interpolation, even if we use linear interpolation for rotations instead of the more expensive "slerp".
However, Gaussians may have the same shape (covariance), even though they differ significantly in their rotation and scaling. 
For instance, a Gaussian that is scaled by $s$ along its $x$-axis will appear the same as a Gaussian that is scaled by $s$ along it's $y$-axis and then rotated by $90^{\circ}$ around its z-axis (Fig.~\ref{fig:interp}(a)). 
As a result, when directly interpolating their properties, an undesired rotation will occur which is visually disturbing.

To avoid this, during hierarchy generation we also perform orientation matching: starting from the root node, we recursively reinterpret the axes of orientation for each child Gaussian such that it minimizes the relative rotation between the child node and its parent, by exhaustive search. Using a non-exhaustive approach, e.g., matching eigenvalues, unnecessary rotations still occur.

\begin{figure*}[!h]
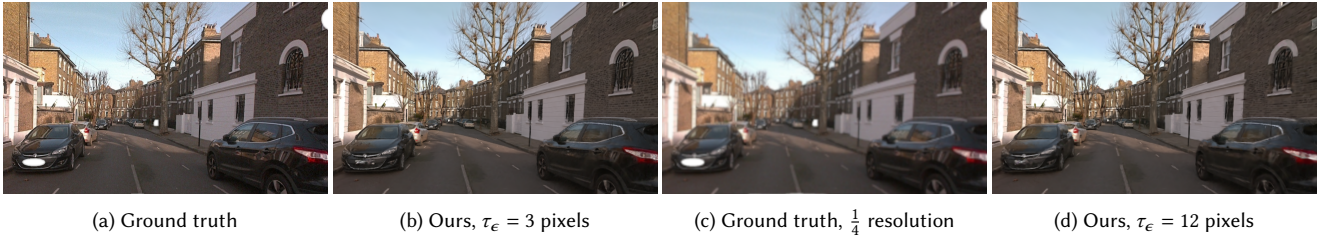

	\begin{subfigure}{0.24\textwidth}
		\includegraphics[width=\textwidth]{figures/comp_hi}
		\caption{Ground truth}
	\end{subfigure}
	\begin{subfigure}{0.24\textwidth}
		\includegraphics[width=\textwidth]{figures/comp_hi_ours}
		\caption{Ours, $\tau_\epsilon = 3$ pixels}
	\end{subfigure}
	\begin{subfigure}{0.24\textwidth}
		\includegraphics[width=\textwidth]{figures/comp_lo}
		\caption{Ground truth, $\frac{1}{4}$ resolution}
	\end{subfigure}
	\begin{subfigure}{0.24\textwidth}
		\includegraphics[width=\textwidth]{figures/comp_lo_ours}
		\caption{Ours, $\tau_\epsilon = 12$ pixels}
	\end{subfigure}
	\caption{(a) Ground truth image and (b) fine-detail setting of our method. In contrast to conventional multi-scale training, we supervise all levels on the full resolution instead of rescaled images (c), which would encourage blur. Our approach preserves sharp high-frequency features at much coarser settings (d).
		\label{fig:simple-hier}
	}
\end{figure*}

Interpolation of opacity also requires careful treatment. At the start of the transition from a parent to the children, the children share all other attributes with the parent that are progressively interpolated to those of each individual child (Fig.~\ref{fig:interp}(b)). However, we need to modify the \emph{falloff} of the child nodes so that the resulting blending of the overlapping Gaussians gives the same result as the parent at the start of the transition.

We seek a blending weight $\alpha'$ for all children such that the resulting blended contribution is equal to that of the parent, at the start of the transition. 
Consider the simple case of two children and an isolated parent node. Just before the point of transition, the blended color of the parent is $\alpha_p c_p$ (Eq.~\ref{eq:alpha}), where $c_p$  is the color of the parent. We need to solve for $\alpha'$ such that:
\begin{equation}
	\alpha_p c_p ~=~ \alpha' c_p + (1-\alpha')\alpha' c_p
\end{equation}
\noindent
so that at the start of transition the blended color of the children is exactly that of the parent. Solving gives:
\begin{equation}
	\label{eq:alphap}
	\alpha'~=~1 - (1 - \alpha_p)^{(\frac{1}{2})}
\end{equation}
\noindent
in the case of two children.
This $\alpha'$ weight is used to blend $\alpha$ from the parent to the children with the same linear interpolation scheme as the other attributes for each child node $i$:
\begin{equation}
	\alpha (t) ~=~ t \alpha_i + (1-t) \alpha'
\end{equation}
With this interpolation scheme, we achieve smooth transitions with our hierarchy.
We illustrate the resulting renderings of our hierarchy for two different target granularities in Fig.~\ref{fig:simple-hier}; please see the supplemental video that illustrates the smooth transitions.

\section{Optimizing and Compacting the Hierarchy}
\label{sec:compact-hier}

The hierarchy is constructed by aggregating geometric primitives; ultimately we need to take appearance into account more explicitly. Since
each intermediate node of the hierarchy is in itself a 3D Gaussian primitive, 
it can be further optimized to improve visual quality. To do this, we need to propagate gradients through the intermediate nodes, introducing a hierarchy that has intermediate nodes we can optimize.
We next explain how this is achieved, together with an additional step to compress the hierarchy.

\begin{figure}[!h]
\includegraphics[width=0.85\linewidth]{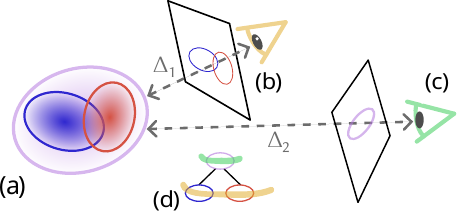}
\caption{
\label{fig:scale-select}
	(a) The purple node contains the red and blue nodes in the hierarchy. The most common understanding of LOD is based on distance $\Delta$: When the viewpoint is close, we descend in the hierarchy (b) while when further away we use the higher-level node (c). Corresponding cuts are illustrated in (d).
}
\end{figure}

Traditionally LOD methods in graphics are used to represent a simplified version of the scene when viewed from afar; this is illustrated in Fig.~\ref{fig:scale-select}. 
Our target granularity achieves this effect by reasoning in terms of projected screen area, expressed with the granularity $\epsilon(n)$ of node $n$, defined previously.

\subsection{Optimizing the Hierarchy}
\label{sec:post-optim}

\begin{figure}[!h]
	\includegraphics[width=0.81\linewidth]{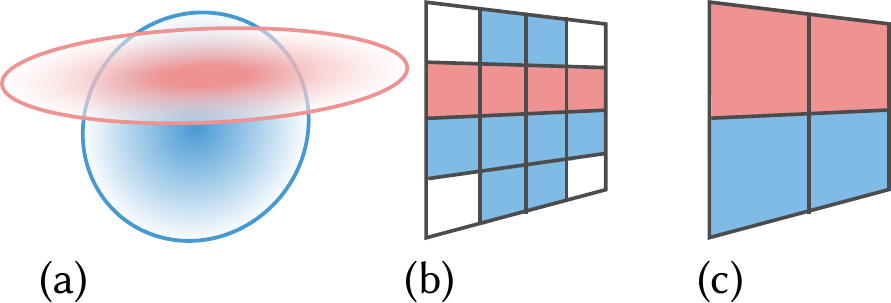}
	\caption{
		\label{fig:gran-vs-res}
		(a) Two nodes in a hierarchy. In (b) we choose a target granularity of 2x2 that results in the cut in (a). If we instead force image resolution and cut to always match (c), we relinquish the ability of large, anisotropic Gaussians to model higher-frequency details.
	}
\end{figure}

Our top-down hierarchy construction results in a data-structure that works well (see Tab.~\ref{tab:comparisons}). 
However, once constructed, the intermediate nodes can be rendered the same way as leaf 3DGS primitives, and thus can be
optimized the same way. As a result, we can 
optimize the intermediate nodes of the hierarchy to improve the visual quality they can represent. 
This poses the question of how to perform this optimization 
between different scales. One solution would be to optimize the hierarchy by randomly selecting an input view and a downsampling factor~\cite{barron2021mip}; a lower resolution directly implies a different target granularity, and the corresponding hierarchy cut. However, this method has drawbacks that we illustrate in Fig.~\ref{fig:simple-hier} and Fig.~\ref{fig:gran-vs-res}: when reducing resolution, there are high-frequency details that cannot be represented.

Instead of downsampling, we always render at full resolution during optimization, and simply choose random target granularities. This has the desired effect of optimizing nodes in many different cuts, while preserving visual details.  This can be particularly useful when we have limited resources, where we can apply a more aggressive LOD strategy while maintaining better visual quality.

Concretely, we load the generated hierarchy and optimize it by randomly selecting one of the training views and a target granularity $\tau_{\epsilon}$ in a given target range at random.
The target range $[\tau_{min}, \tau_{max})$, and the training views chosen define the corresponding cuts selected.
To achieve good quality sampling of all levels of the hierarchy, we sample the target granularities $\tau_{\epsilon}$ given a canonical random variable $\xi \in \left[0, 1\right)$ using $\tau_{\epsilon} = \tau_{max}^\xi~\tau_{min}^{1-\xi}$.

Both cut selection and interpolation (Sec.~\ref{sec:methodlod}) are used during this optimization.
To optimize child and parent nodes simultaneously with smooth switching enabled, we must propagate gradients correctly to two layers of the hierarchy, optimizing child and parent nodes simultaneously.
This requires gradients to propagate through the interpolation weights and the expression of $\alpha'$ in Eq.~\ref{eq:alphap}. 
By optimizing multiple levels of the hierarchy, we tackle a more complex setting than 3DGS: Optimizing higher LOD levels may also degrade the quality of the leaves due to interpolation. To avoid this, we do not change the leaf nodes during optimization.

Note that, as observed by recent work~\cite{geiger}, the original 3DGS method does not handle antialiasing properly; since hierarchies operate at different scales, correct anti-aliasing is required. We thus use the EWA filtering solution of Yu et al.~\shortcite{geiger}.

\subsection{Compacting the Hierarchy}
\label{sec:compact}

The hierarchy adds some overhead in terms of memory. More importantly, for the hierarchy optimization itself
we want to avoid parent-child settings where the parent's size is only marginally larger than the children's. Otherwise, such nodes might be selected rarely and not be properly optimized during training. 
To avoid this we sparsify the generated tree structure.  

We begin by marking all leaf nodes --- the output of 3DGS optimization --- as relevant, i.e., they should not be removed from the tree.
Next, we find the union of cuts in the tree (according to Sec.~\ref{sec:select}) over all training views for the minimum target granularity $\tau_{min} = 3$ pixels (the minimum extent of a primitive 3DGS primitive due to low-pass filtering).
We then find the bottom-most nodes in this union, which again yields a cut. 
These nodes are considered to be the highest-detail nodes that are relevant for the selected granularity. 
All nodes between them and already marked nodes are removed from the tree. We then raise the target granularity by $2\times$ and repeat this process until $\tau_{max}$, half the image's resolution, is reached. 
Note that this may result in nodes with $K$ children, in which case Eq.~\ref{eq:alphap} generalizes to:
$\alpha'~=~1 - (1 - \alpha_p)^{(\frac{1}{K})}$.

\section{Large Scene Training}
\label{sec:methodlarge}

\begin{figure}
	\begin{tabular}{cc}
		{\includegraphics[width=0.45\columnwidth]{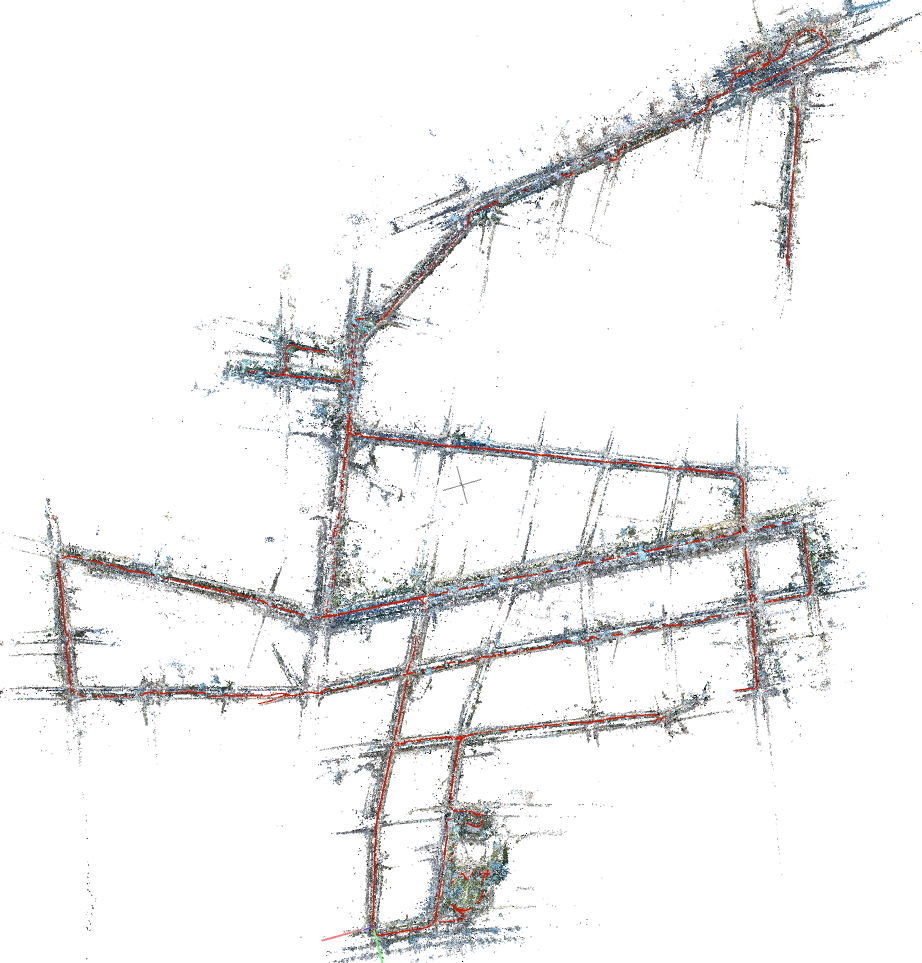}} &
		{\includegraphics[width=0.45\columnwidth]{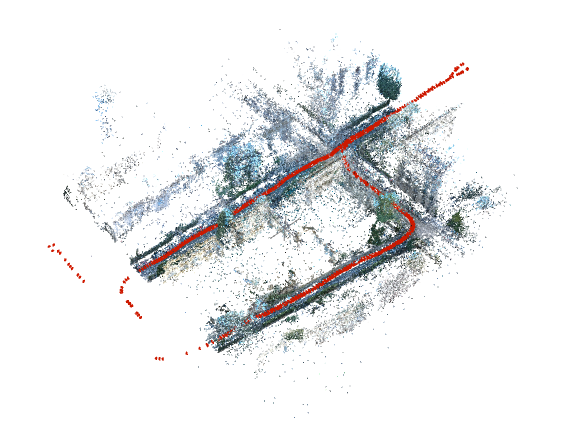}} \\
	\end{tabular}
	\caption{
 \label{fig:chunks}
Left: The \textsc{BigCity} capture with the calibrated cameras and SfM points (28\,145 images total for an area of approximately 2$\times$2 km). Right: An example chunk of this dataset, supervised by 1\,301 images.
}
\end{figure}

We can now build efficient hierarchies of 3D Gaussians; these are indispensable for the processing of very large scenes, since parts of the scene that will be seen from far away can be rendered at coarser levels of the hierarchy.
To train large scenes we build on common computer graphics methodologies for real-time rendering of large data~\cite{luebke2003level}.
In particular, we introduce a divide-and-conquer approach by subdividing large scenes into \emph{chunks} (Fig.~\ref{fig:chunks}).

We define chunk size as 50$\times$50\,m for scenes captured walking to 100$\times$100\,m for scenes captured with a vehicle. 
While limited in size, these chunks are still larger than those treated by the original 3DGS approach. In addition, the capture style of such scenes is necessarily significantly sparser than those expected by most radiance field solutions.
We thus adapt the optimization of the 3D Gaussians to account for these differences.

Our goal is to allow parallel processing of individual chunks, allowing the processing of large scenes in reasonable wall-clock time given sufficient computational resources. Once the individual chunks have been processed, we need a \emph{consolidation} step to handle potential inconsistencies between the individual chunks.

\subsection{Coarse Initialization and Chunk Subdivision}

We first calibrate cameras of the entire dataset; we discuss our solution to this engineering challenge in Sec.~\ref{sec:impl-data} and in the Appendix. 
To allow consistent training of all chunks, we need to provide
a basic \emph{scaffold} and \emph{skybox} for all ensuing steps. We do this by running a very coarse initial optimization on the full dataset. 
Specifically, we initiate a default 3DGS optimization of the entire scene, using the 
available SfM points and add an auxiliary \emph{skybox} (see Sec.~\ref{sec:chunk-train}). 
Additionally, we disable densification and the position of primitives is not optimized during this step, since the SfM points are well placed.
This coarse model serves as a minimal basis for providing backdrop details, i.e., parts of the scene outside a given chunk. 
In the case of extremely large scenes where 
storage of the SfM points would exceed (V-)RAM capacities, the coarse optimization itself could be broken into multiple steps, with intermediate, partial results being streamed out to disk.
We split our scenes into chunks that are large enough to establish sufficient context for common real-world elements, including cars, buildings and monuments. For each chunk, we select all cameras that are inside the bounds of the chunk, or are within $2\times$ the chunk bounds and have more than 50 SfM points within the chunk bounds.

\subsection{Chunk-scale Training}
\label{sec:chunk-train}

We train each chunk independently; the result is then used to create a hierarchy (Sec.~\ref{sec:methodlod}) for each chunk, followed by the optimization and consolidation steps (Sec.~\ref{sec:consolidation}).
A typical chunk in large datasets differs significantly from the data used in the original 3DGS setting. In particular, the extent of the scene is significantly larger than the small scenes of datasets such as Mip-NeRF 360~\cite{barron2022mipnerf360} or Deep Blending~\cite{hedman2018deep}. Also the capture density is much lower, and is not ``object-centric''. This makes it harder to optimize, since the space of rays is not covered uniformly~\cite{kopanas2023improving}. The data also contain exposure changes, humans and moving objects (cars, bicycles etc.) that need to be removed and ignored in the optimization. 

We define a \emph{skybox} that surrounds the extent of the scene, i.e., 100\,000 3DGS primitives on a sphere $10\times$ the diameter of the scene,
to capture the effect of the sky. 
We load the coarse optimization of the scene that will be used for all content outside the chunk; it also prevents each chunk from creating inconsistent content for the sky. We train the content inside the chunk using the optimization of 3DGS~\cite{kerbl3Dgaussians}, using correct anti-aliasing~\cite{geiger}. 
We perform a small, temporary optimization of the coarse environment and the \emph{skybox} outside the chunk. Specifically we only optimize opacity and SH coefficients.

The original 3DGS optimization collects statistics to decide whether or not Gaussians should be densified at regular intervals. 
Specifically, the densification policy is based on the \emph{mean} of the screen-space positional gradients over a fixed number of iterations.
There are two main problems in the context of unbounded scenes. 
First, this policy rarely discourages Gaussians  from densifying, regardless of whether they are fine enough to model local detail. 
Second, data sets with sparse, scattered cameras that observe separate portions of the scene (as is, e.g., the case for urban drive-through captures) result in a much lower tendency to densify overall. 
We address both of these issues by changing the densification policy to consider the \emph{maximum} of the observed screen-space gradient rather than its mean, that is no longer reliable in the context of sparse capture.

The sparse camera captures that we have do not provide enough information for good quality reconstruction, e.g., for the street in urban driving scenarios. We perform monocular depth prediction, scale and shift the depth based on the SfM points and use it to regularize the optimization. This results in improvement in visual quality, especially for the road (see Sec.~\ref{sec:eval}). We provide the details of this process in Appendix~\ref{sec:reg}.

\subsection{Chunk Consolidation and Rendering}
\label{sec:consolidation}

Each chunk starts with the SfM points from the per-chunk refinement from COLMAP (Sec.~\ref{sec:dataset}), and the 3D Gaussians contained in the \emph{scaffold} that are in its neighboring chunks. The resulting 3D Gaussians associated with a chunk and its hierarchy are thus sometimes outside the chunk itself. During the consolidation phase, if a primitive associated to---but outside---chunk $i$ is closer to another chunk $j \neq i$, it is deleted. The consolidation also creates a global hierarchy with a root node for the entire scene. 
Rendering is performed by setting a granularity threshold, and finding the corresponding cut. We update the cut every 2 frames to add details by transferring nodes to the GPU from CPU RAM, and run cleanup every 100 frames.

\section{Implementation, Capture and Preprocessing}
\label{sec:impl-data}

We next provide details on implementation and on dataset capture and preprocessing.

\subsection{Implementation}

We implemented our method on top of the original 3DGS implementation in C++ and Python/PyTorch, modifying the SIBR~\cite{sibr2020} viewer for fast rendering. We will provide all the source code and the data on publication of our paper, including all scripts for dataset preprocessing (\MYhref{https://repo-sam.inria.fr/fungraph/hierarchical-3d-gaussians/}{see project page}). 
We use auto differentation available in PyTorch for the hierarchy optimization, except for the gradients of Eq.~\ref{eq:alphap} which we derive manually.
Also, during the optimization of the hierarchy, we avoid optimizing leaf nodes by using the stop gradient operator on all Gaussians corresponding to leaf nodes in the hierarchy, effectively freezing their attributes.

\begin{figure}[!h]
\includegraphics[width=\linewidth]{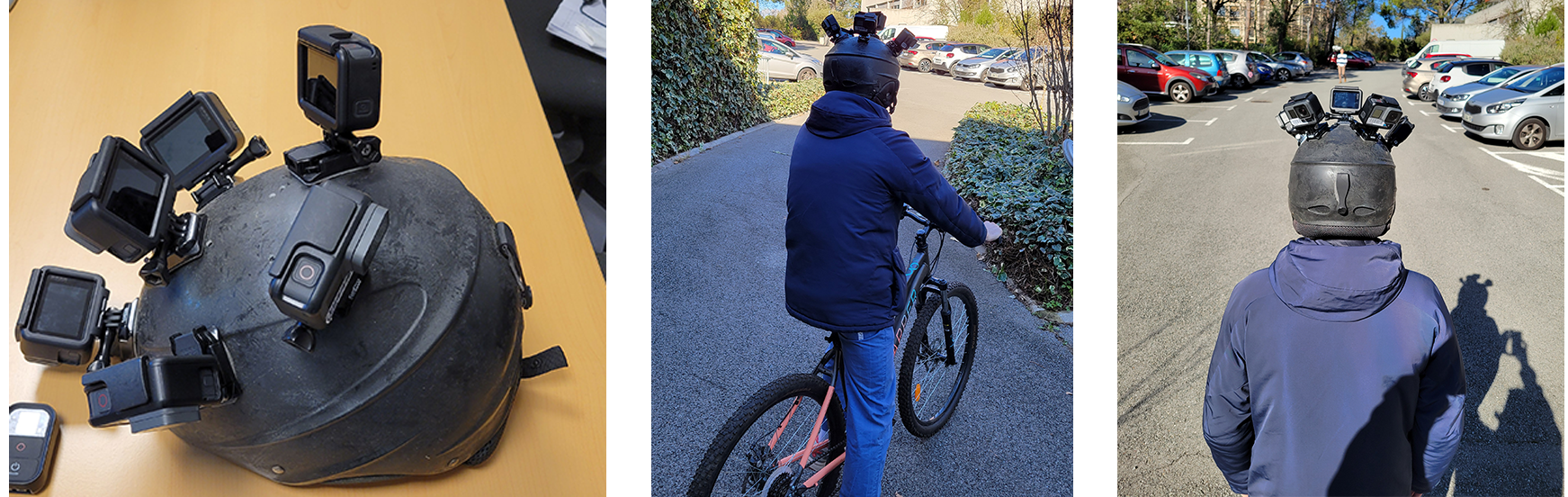}
\caption{
\label{fig:helmet-bike}
Left: Our 6-GoPro HERO6 camera helmet rig. Middle: We performed captures on a bicycle at 6--7\,km/h and (Right) on foot.
}
\end{figure}

\subsection{Dataset Capture and Preprocessing}
\label{sec:dataset}

We captured three outdoors scenes: \textsc{SmallCity}, \textsc{BigCity}, \textsc{Campus}. We list the statistics of each scene in Tab.~\ref{tab:captures} as well as those of an additional scene provided by Wayve.

\begin{table}
\caption{
\label{tab:captures}
Statistics of the three scenes we captured and the one provided by Wayve.
The number of images is given after registration and cleanup.
The area is the area where our chunks are defined.  
}
\begin{tabular}{l|c|c|c|c}
Scene & \#Images & Distance & Area & \#Cameras \\
\hline 
\textsc{SmallCity} & 5\,822 & 450\,m & 40\,000\,m$^2$& 6  \\ 
\textsc{Campus} & 22\,042 & 1.6\,km & 55\,000\,m$^2$& 5  \\ 
\textsc{BigCity} & 38\,235 & 7\,km & 530\,000\,m$^2$& 6  \\ 
\hline
\textsc{Wayve} & 11\,520 & 1\,km & 110\,000\,m$^2$& 6  \\ 
\end{tabular}
\end{table}

\mparagraph{Capture.}
For our captures, we used a bicycle helmet on which we mounted 6 cameras (5 for the \textsc{Campus} dataset). We use GoPro HERO6 Black cameras
(see Fig.~\ref{fig:helmet-bike}), set to Linear FoV and 1444$\times$1080 resolution in timelapse mode with a 0.5\,s step. We performed \textsc{SmallCity} and \textsc{BigCity} captures on a bicycle, riding at around 6--7km/h, while \textsc{Campus} was captured on foot wearing the helmet.

\mparagraph{Pose estimation.}
Pose estimation is a major challenge with the number of cameras we treat. Our datasets have between 5\,800 and 40\,000 photographs. We use COLMAP with customized parameter settings, the hierarchical mapper and an additional per-chunk bundle adjustment to achieve reasonable processing times. We provide details in Appendix~\ref{sec:pose}.

\mparagraph{Dataset Processing.}
We correct for exposure, in a similar spirit to others~\cite{martinbrualla2020nerfw,mueller2022instant}, by optimizing an exposure correction per image. Finally, we remove moving objects by running a CNN-based segmentation on cars, bicycles etc, and determining whether they have corresponding SfM points. We also remove all humans and license plates. Details on all steps are presented in the Appendix.

\subsection{Hierarchy Optimization}
Several intermediate nodes will generate $\alpha$ values $>1$, required for 
merged Gaussians to represent more opaque-appearing primitives with a delayed fall-off (Sec.~\ref{sec:methodlod}).
However, in the presence of these nodes, we can no longer apply the original 3DGS exponential activation for opacity during training.
Instead, we use an absolute value activation function for post optimization.  Since the $\alpha$-blending values are already internally clamped to $0.99$ in the rasterization, no additional changes are necessary to the 3DGS forward routine.
However, for robustness, we must account for this clamping step and zero a Gaussian's opacity gradients whenever it occurs.

\begin{figure*}
	\setlength{\tabcolsep}{2pt}
	\begin{tabular}{cccccc}
		& Ground Truth & Ours opt ($\tau_2, 6$ pixels) & Mip-NeRF360 & Instant-NGP & F2-NeRF \\
		\multirow{2}{*}{\rotatebox[origin=c]{90}{\textsc{Wayve}}}
		& \includegraphics[width=.185\linewidth]{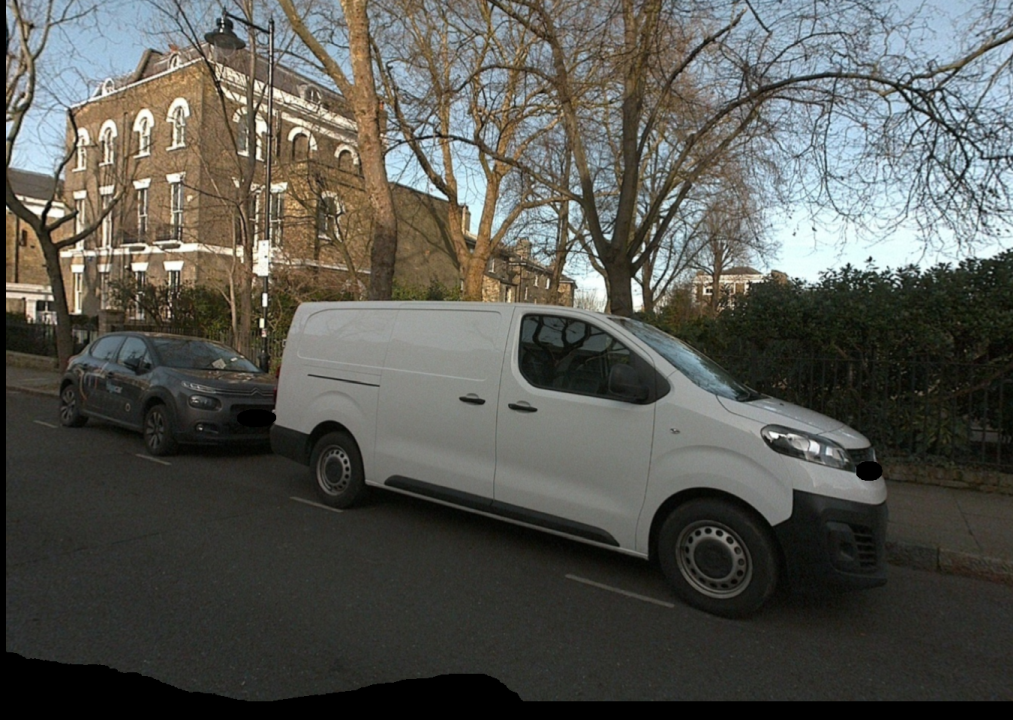} &
		\includegraphics[width=.185\linewidth]{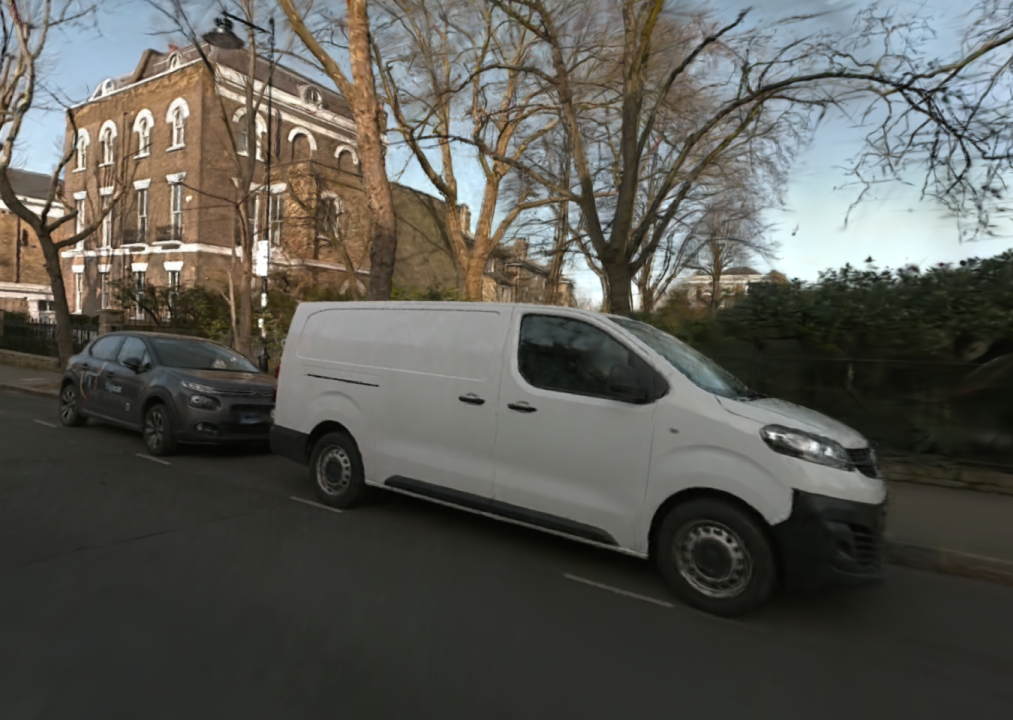} &
		\includegraphics[width=.185\linewidth]{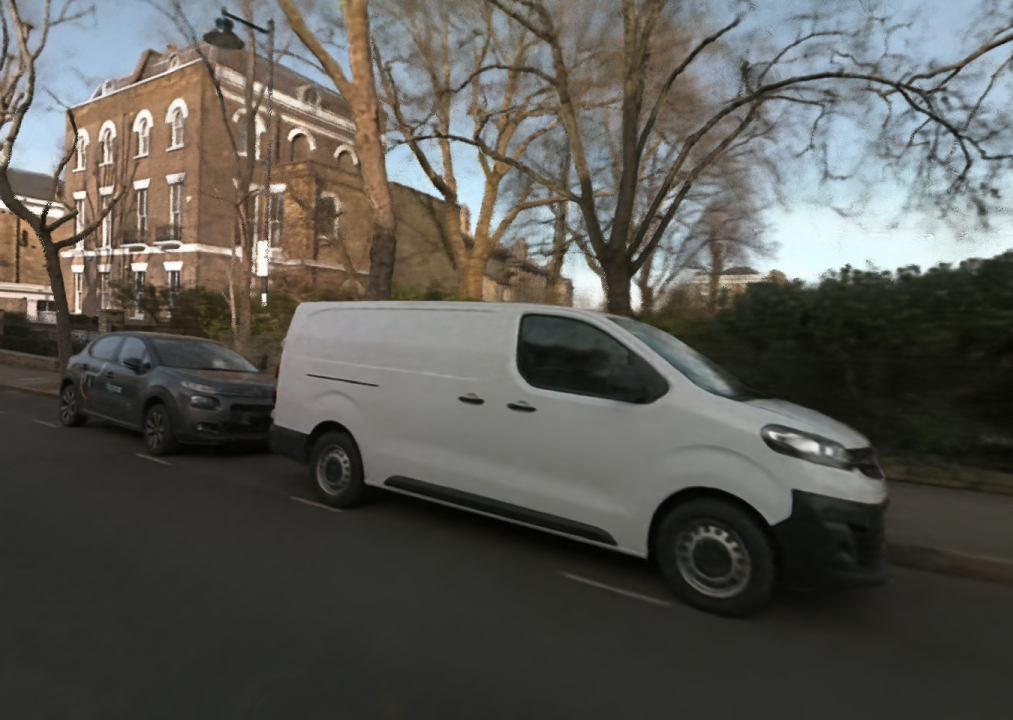} &
		\includegraphics[width=.185\linewidth]{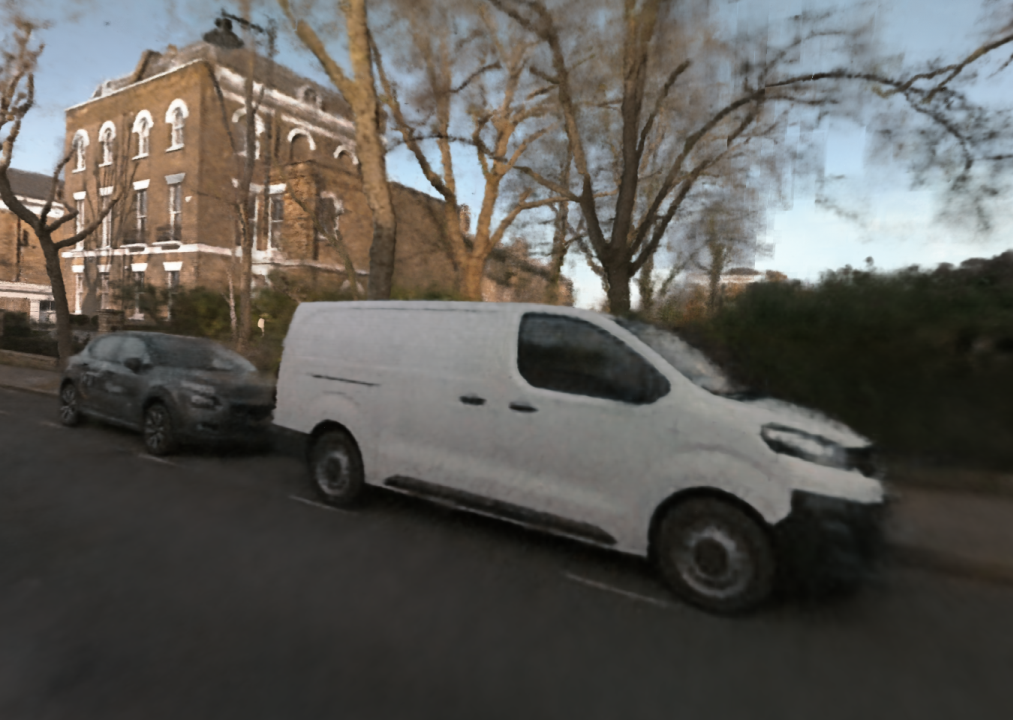} &
		\includegraphics[width=.185\linewidth]{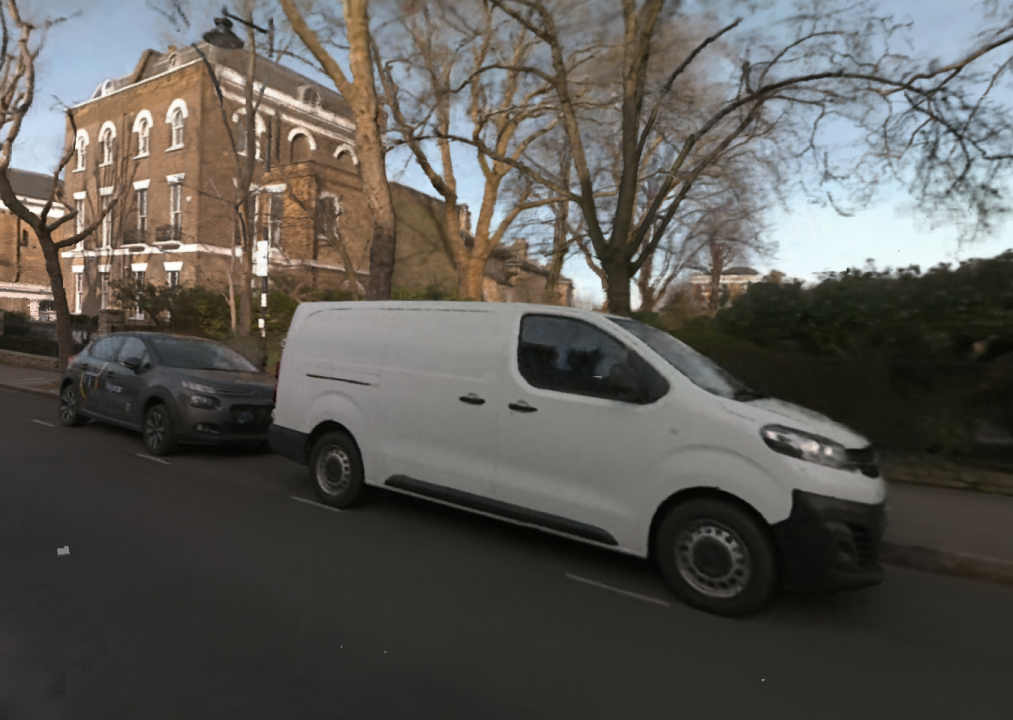} 
		\\
		& \includegraphics[width=.185\linewidth]{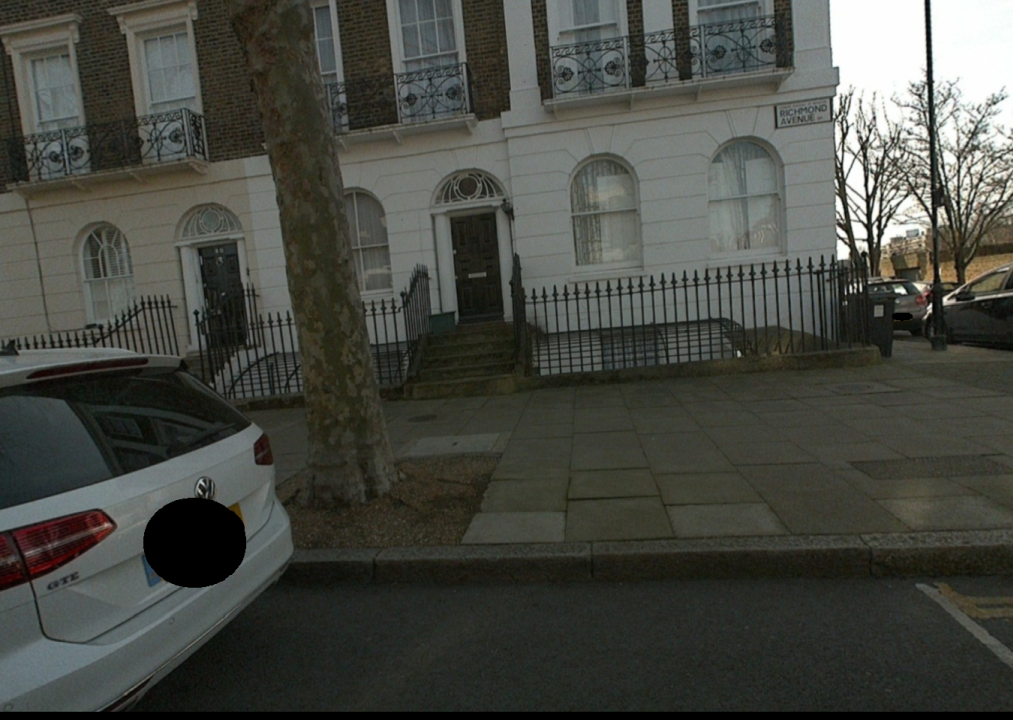} &
		\includegraphics[width=.185\linewidth]{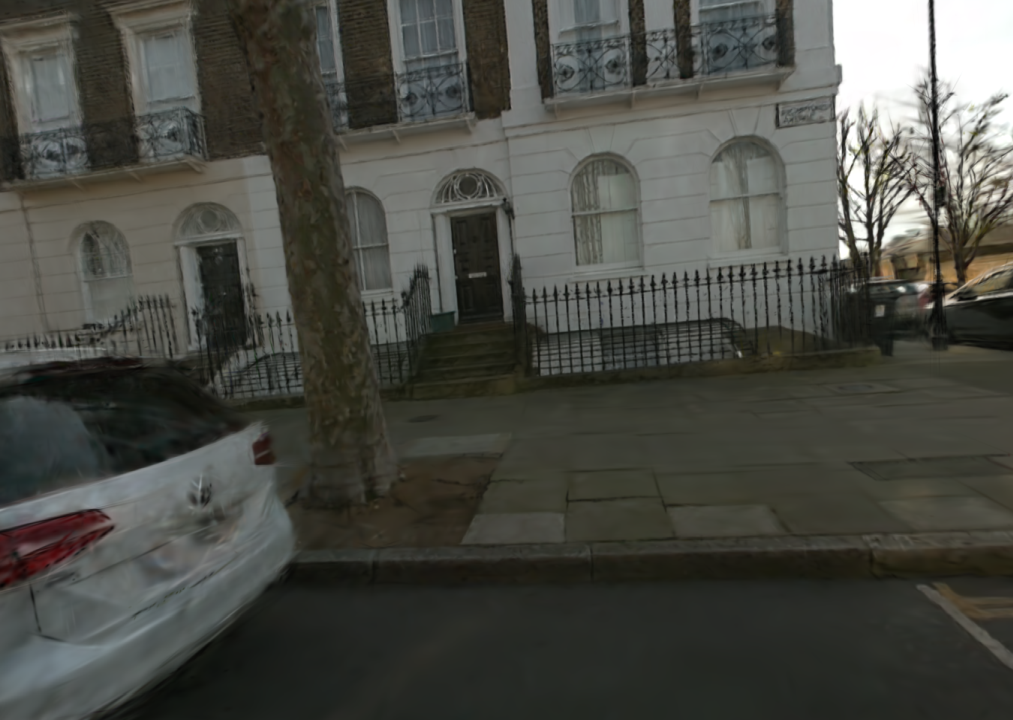} &
		\includegraphics[width=.185\linewidth]{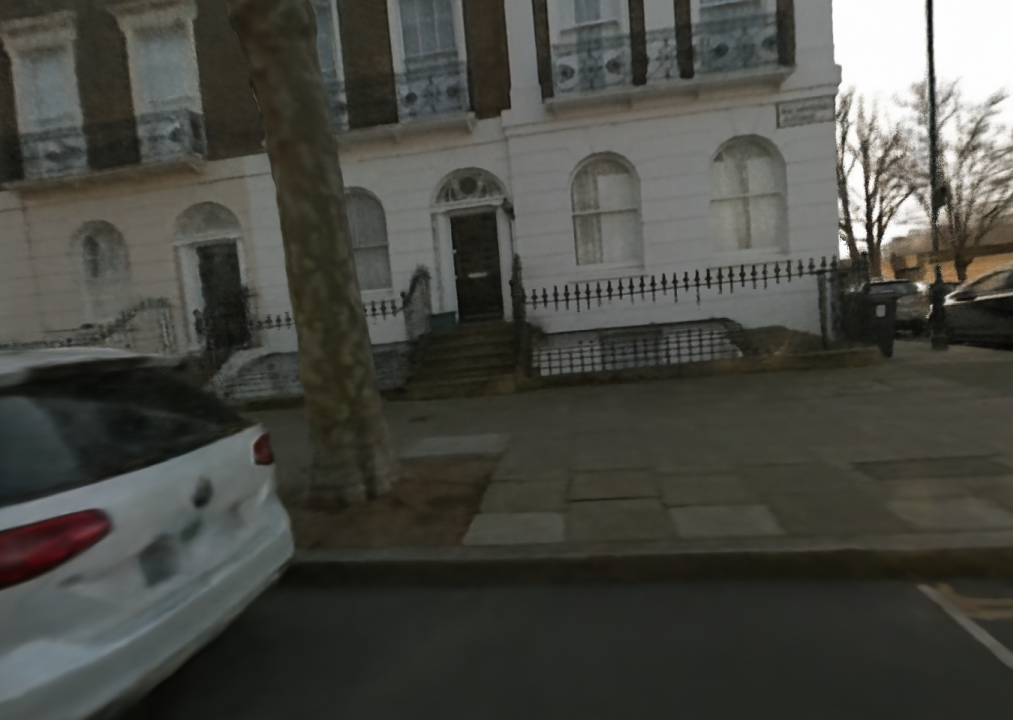} &
		\includegraphics[width=.185\linewidth]{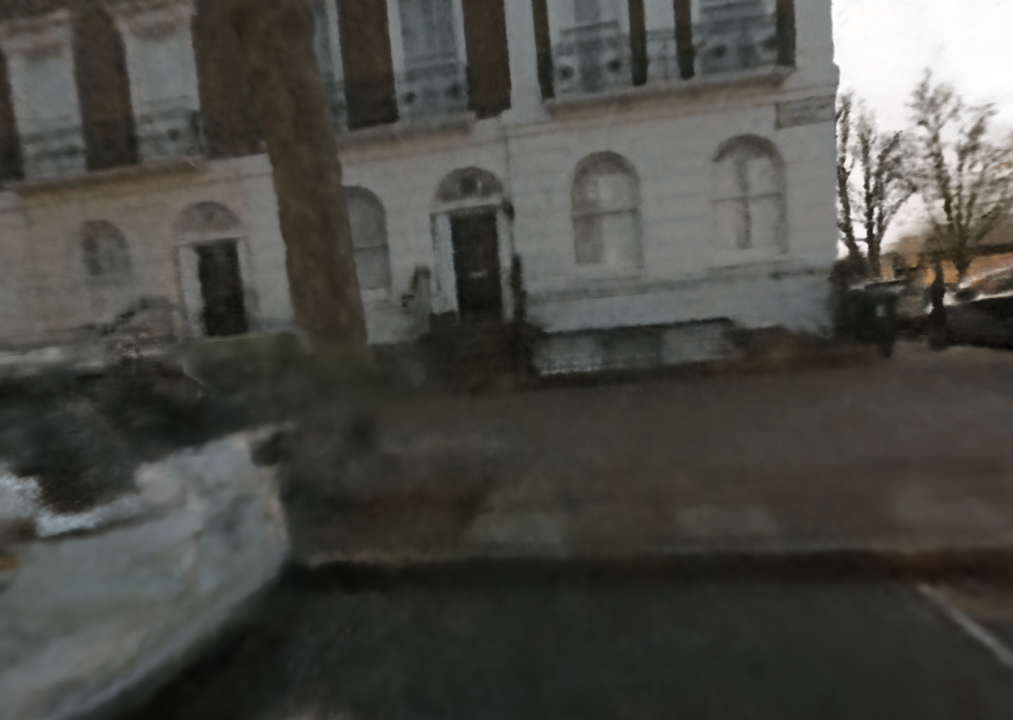} &
		\includegraphics[width=.185\linewidth]{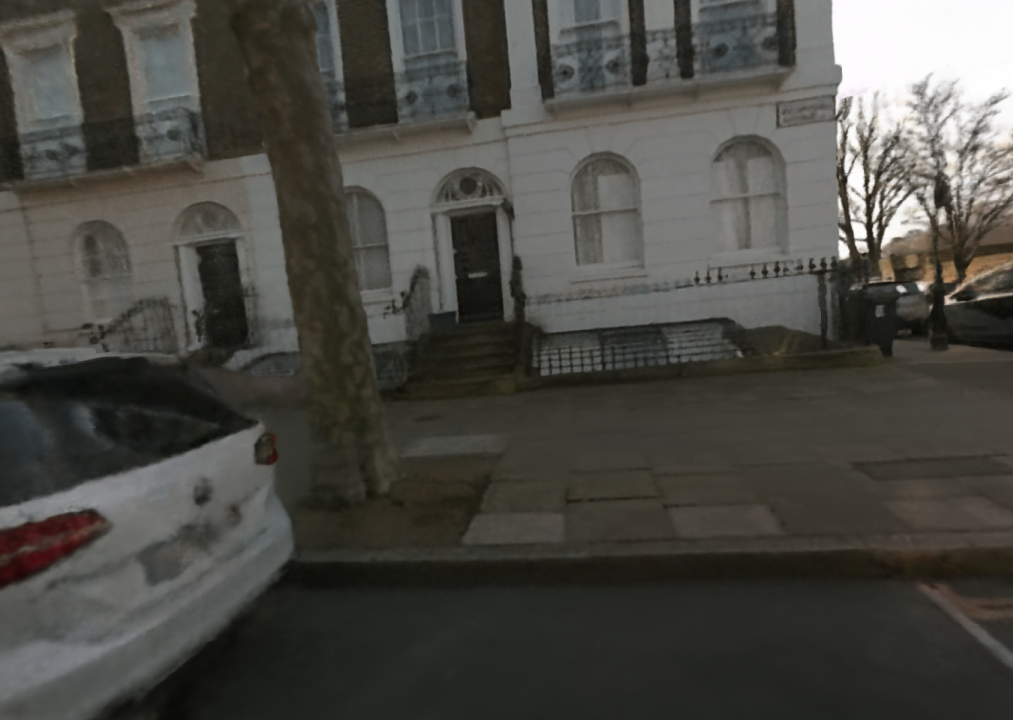} 
		\\
		\multirow{2}{*}{\rotatebox[origin=c]{90}{\textsc{SmallCity}}}
		& 	\includegraphics[width=.185\linewidth]{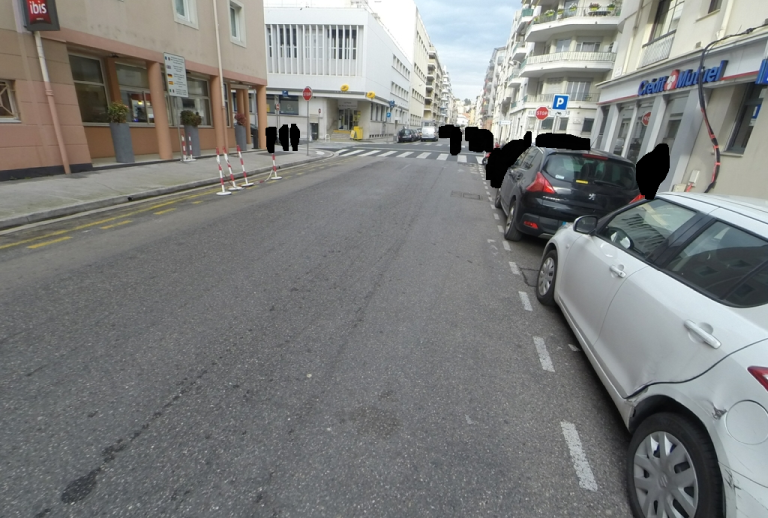} &
		\includegraphics[width=.185\linewidth]{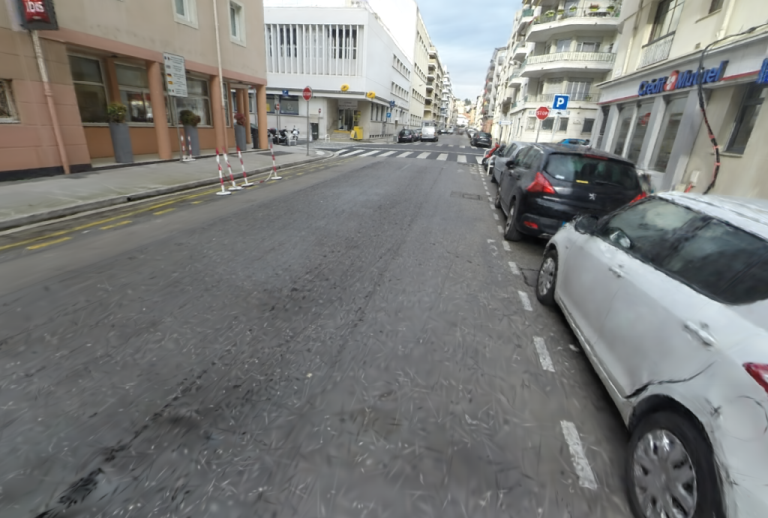} &
		\includegraphics[width=.185\linewidth]{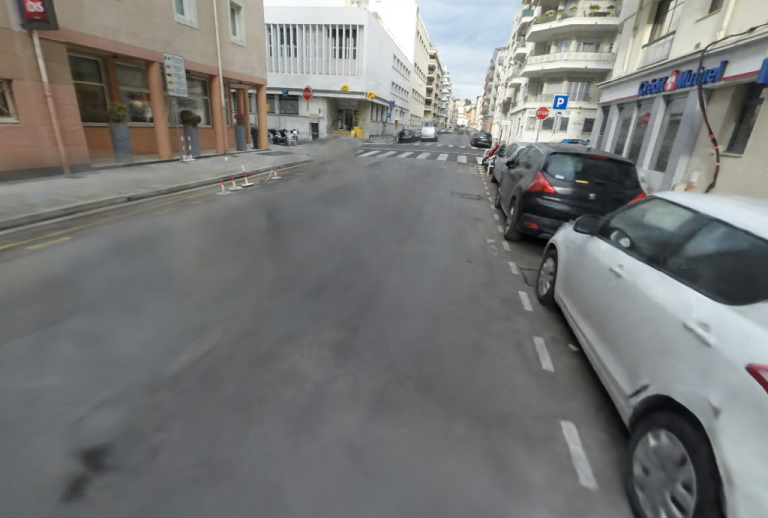} &
		\includegraphics[width=.185\linewidth]{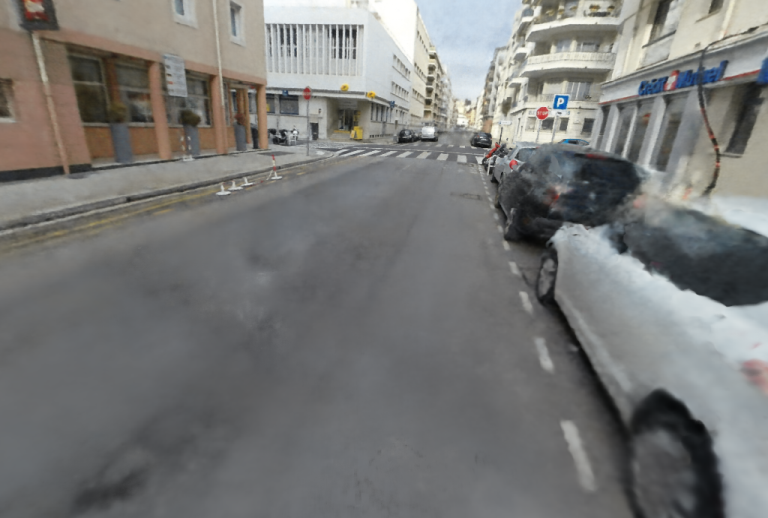} &
		\includegraphics[width=.185\linewidth]{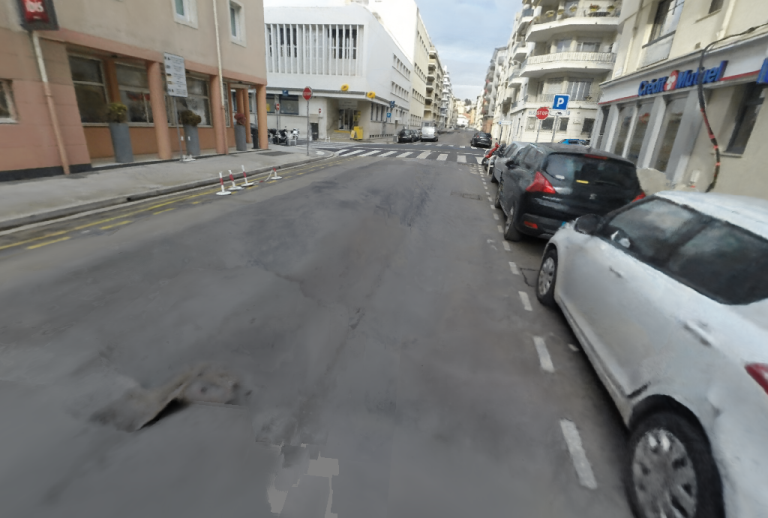} 
		\\
		& \includegraphics[width=.185\linewidth]{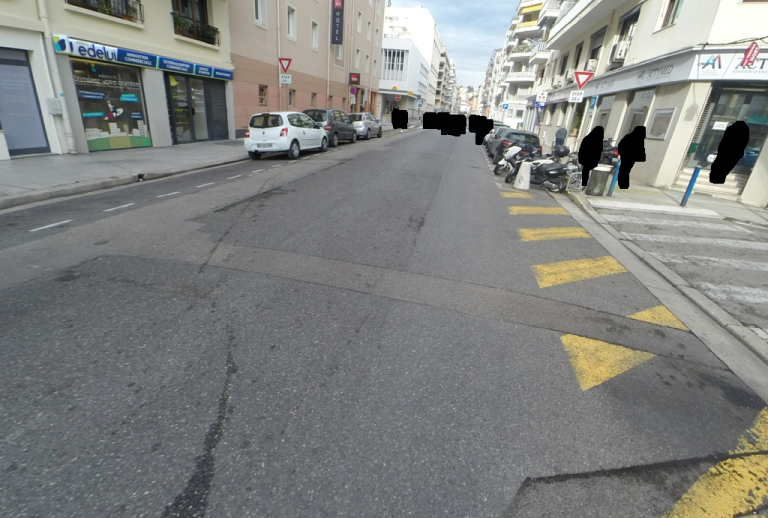} &
		\includegraphics[width=.185\linewidth]{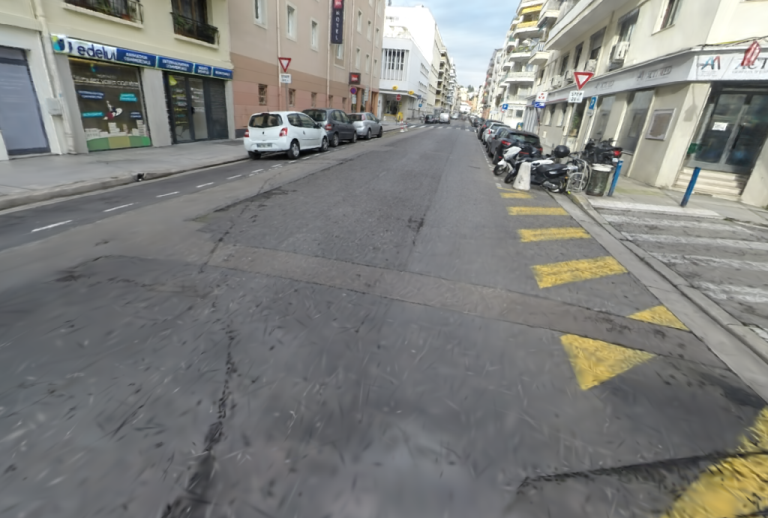} &
		\includegraphics[width=.185\linewidth]{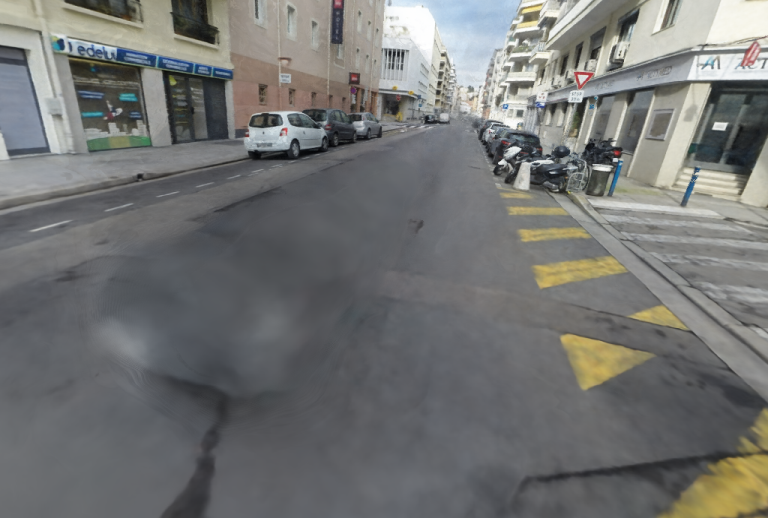} &
		\includegraphics[width=.185\linewidth]{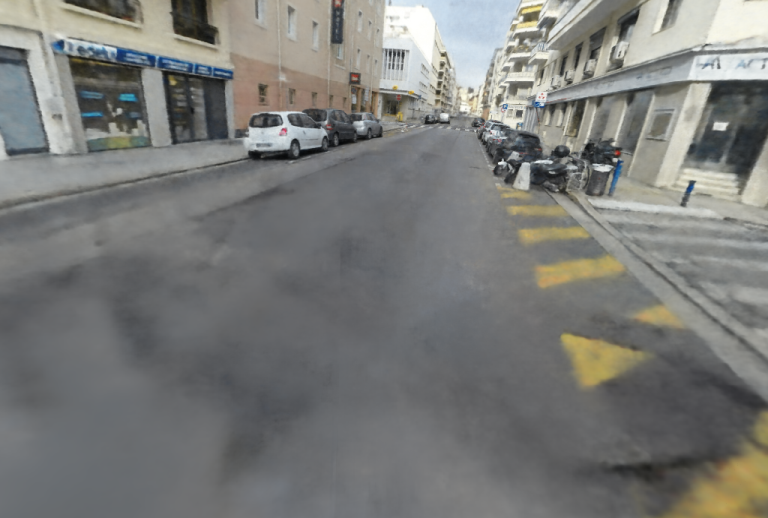} &
		\includegraphics[width=.185\linewidth]{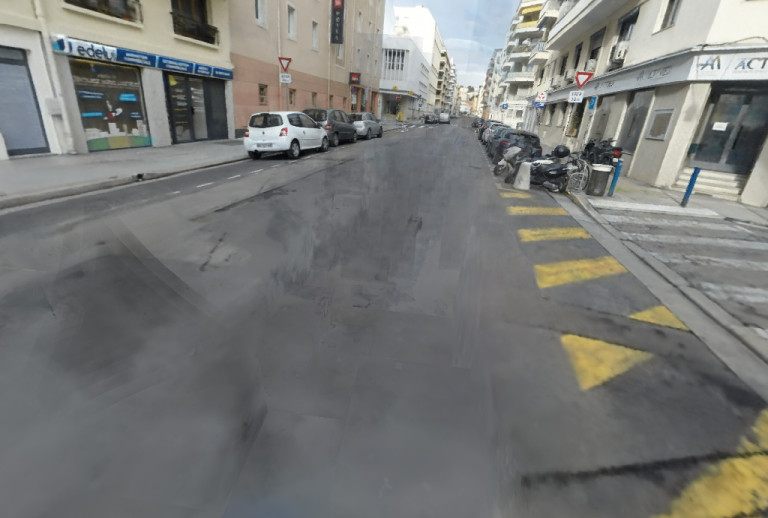} 
		\\
		\multirow{2}{*}{\rotatebox[origin=c]{90}{\textsc{Campus}}}& \includegraphics[width=.185\linewidth]{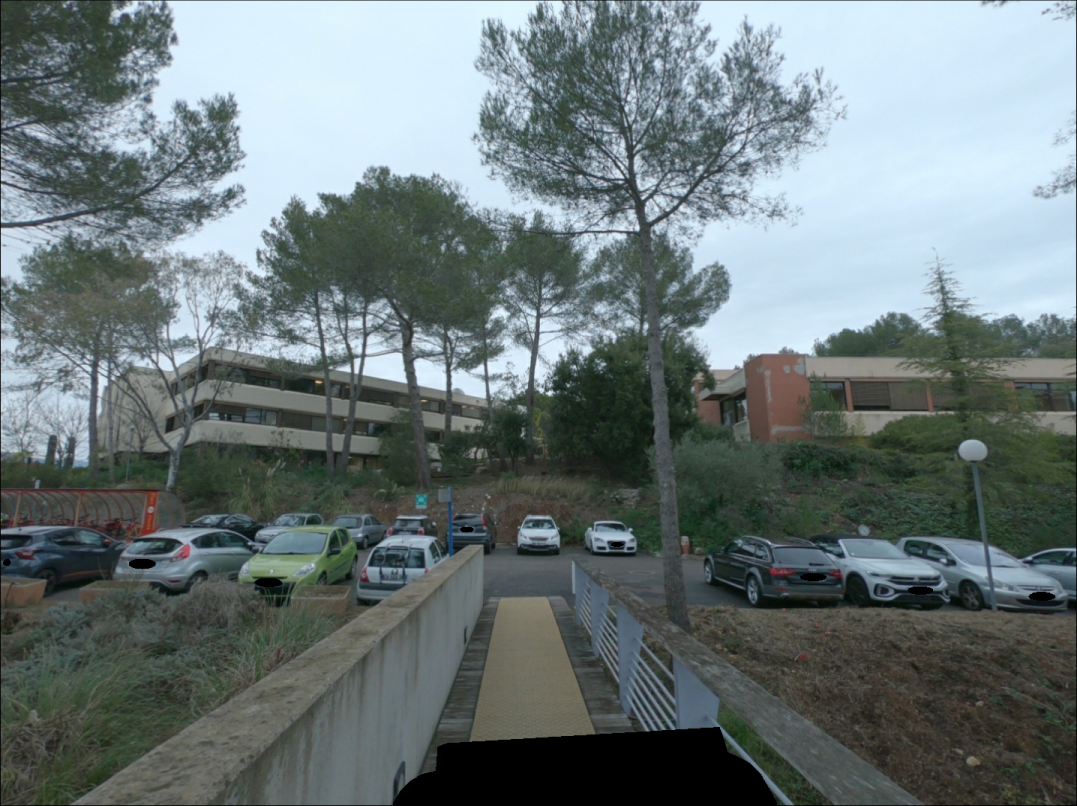} &
		\includegraphics[width=.185\linewidth]{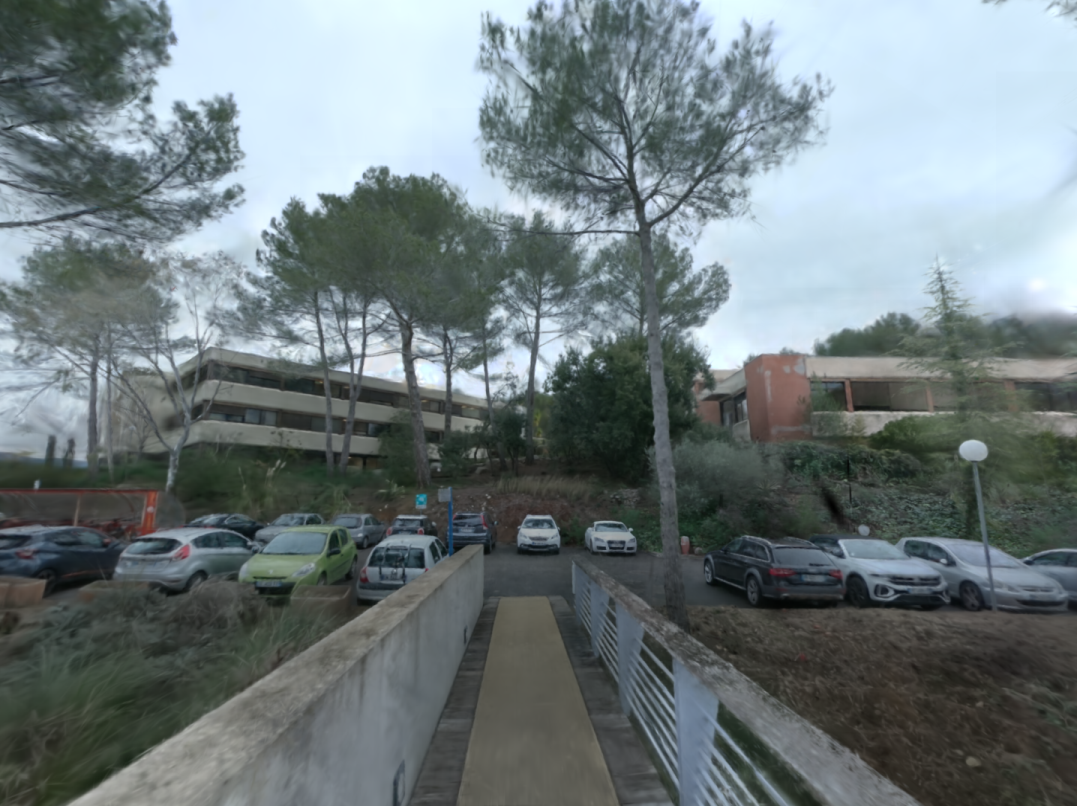} &
		\includegraphics[width=.185\linewidth]{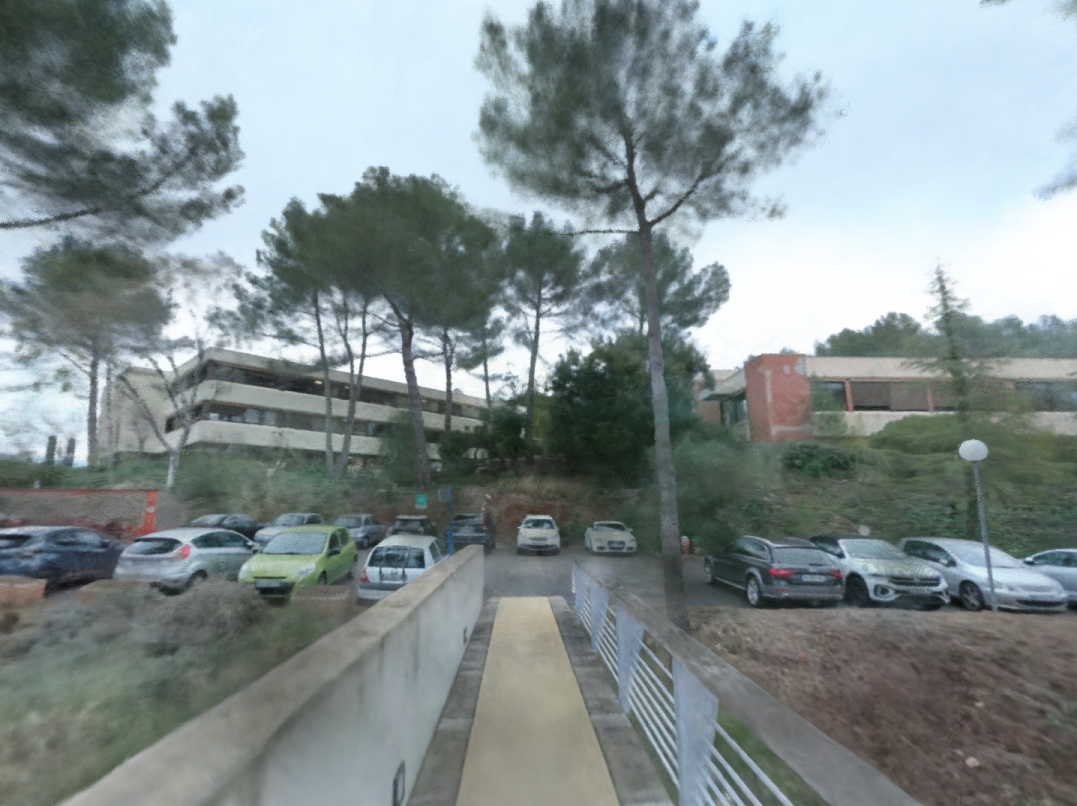} &
		\includegraphics[width=.185\linewidth]{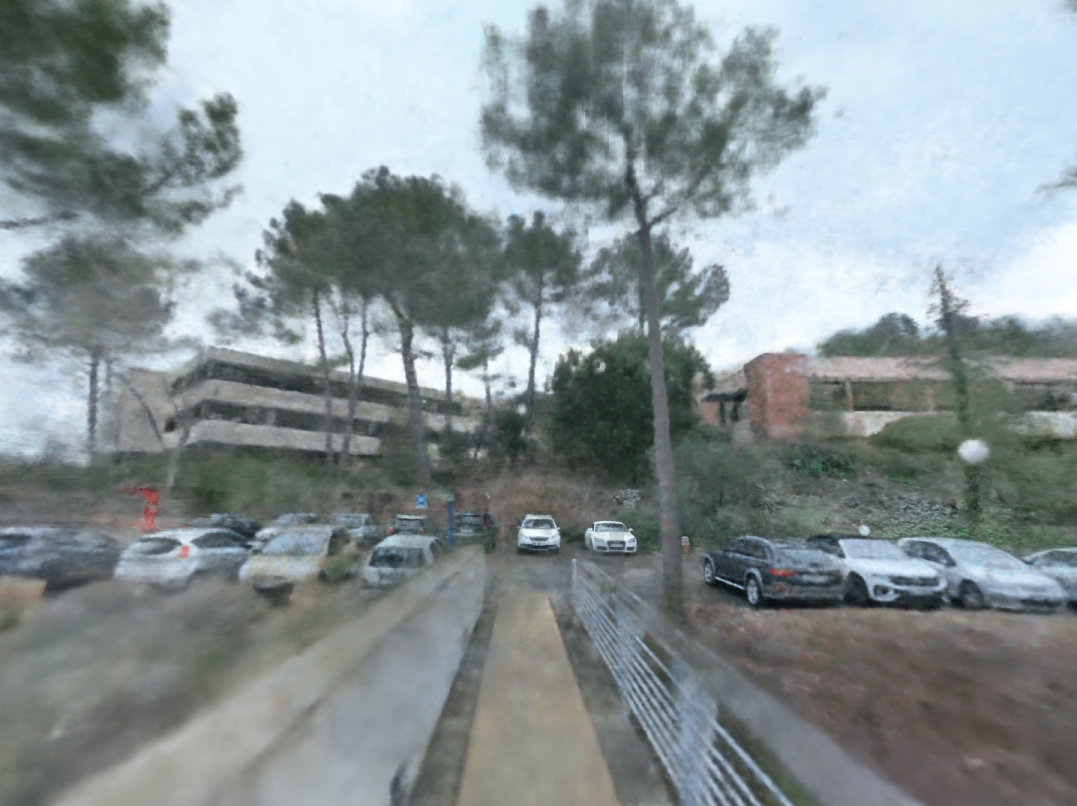} &
		\includegraphics[width=.185\linewidth]{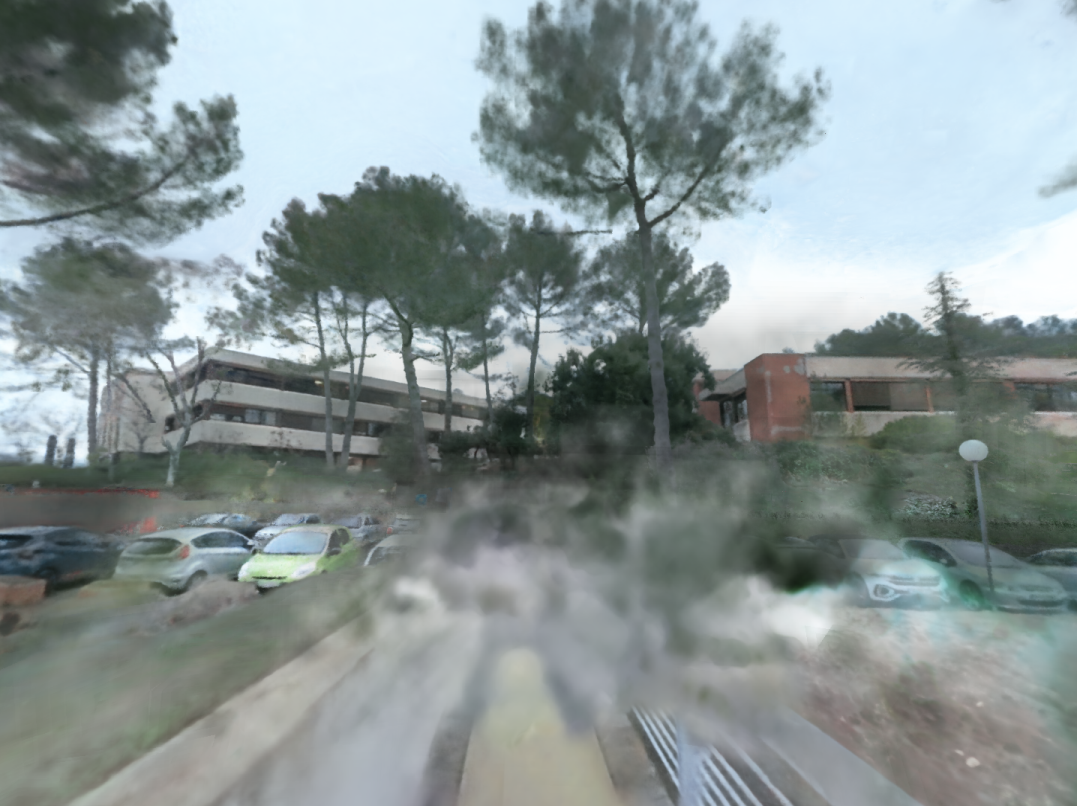} 
		\\
		& \includegraphics[width=.185\linewidth]{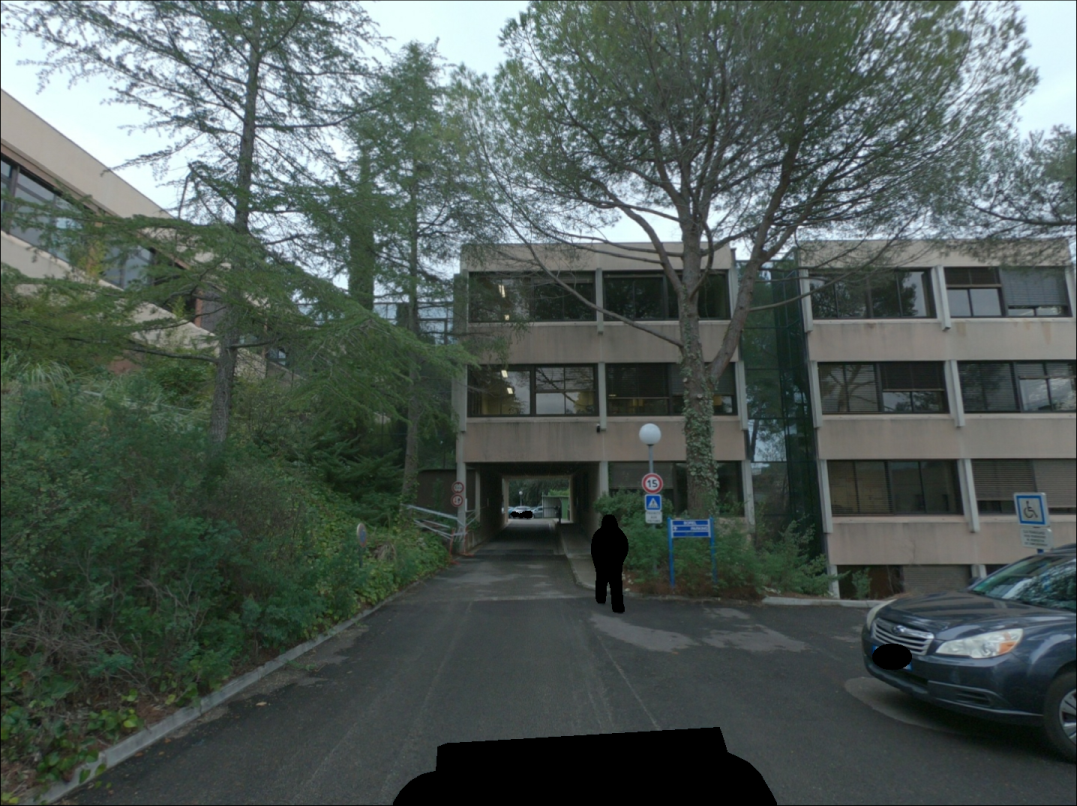} &
		\includegraphics[width=.185\linewidth]{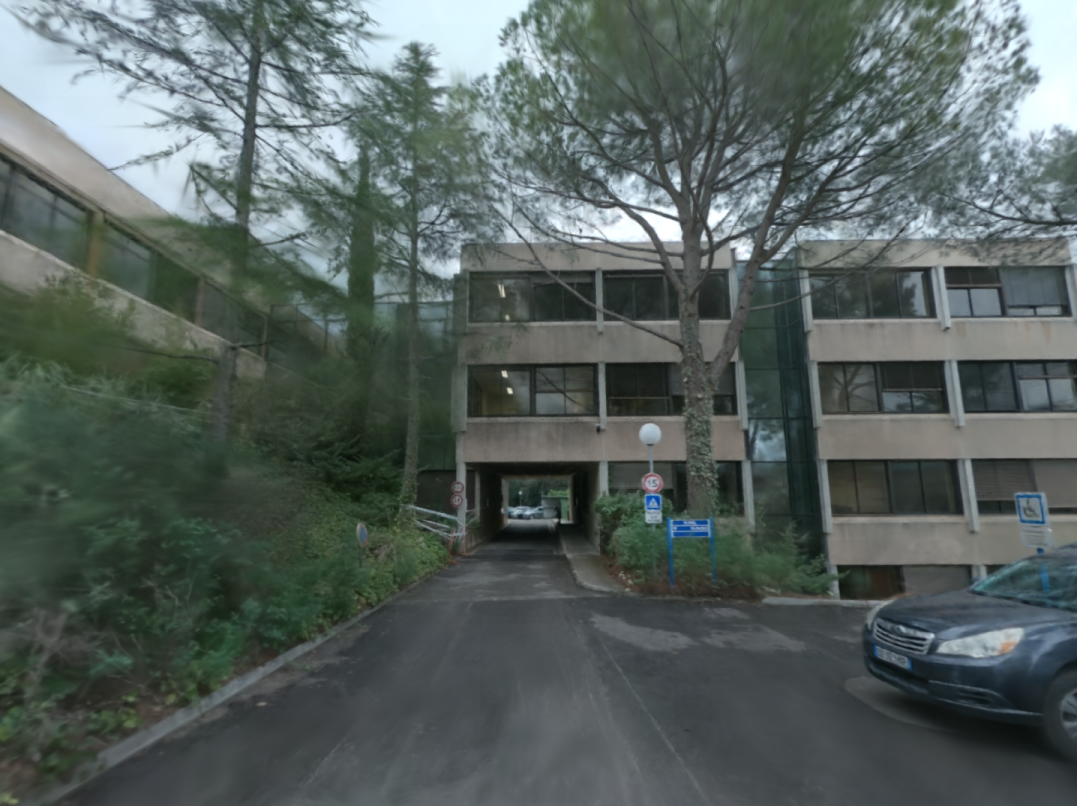} &
		\includegraphics[width=.185\linewidth]{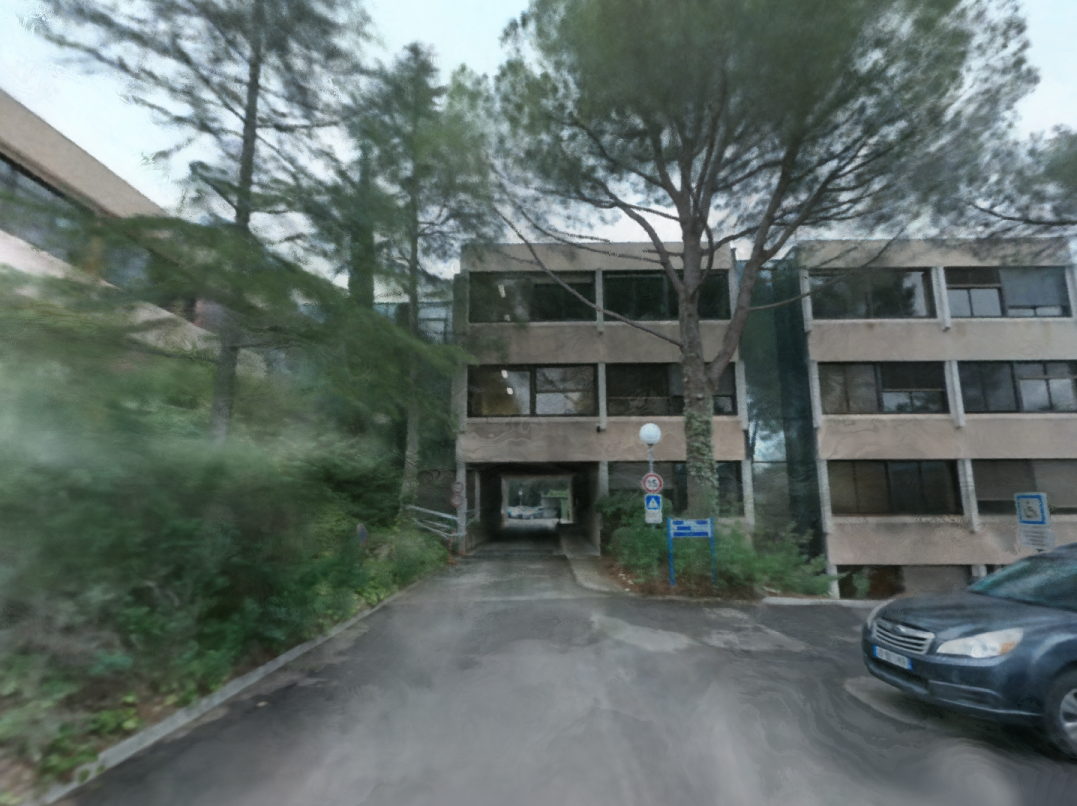} &
		\includegraphics[width=.185\linewidth]{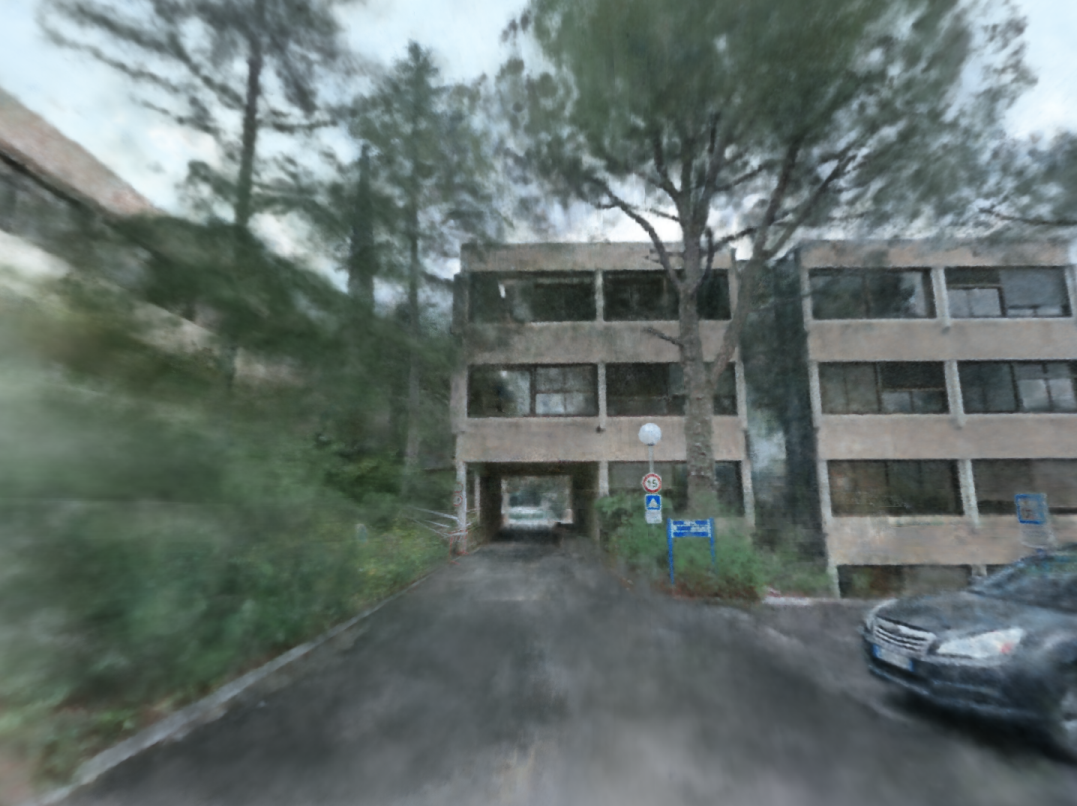} &
		\includegraphics[width=.185\linewidth]{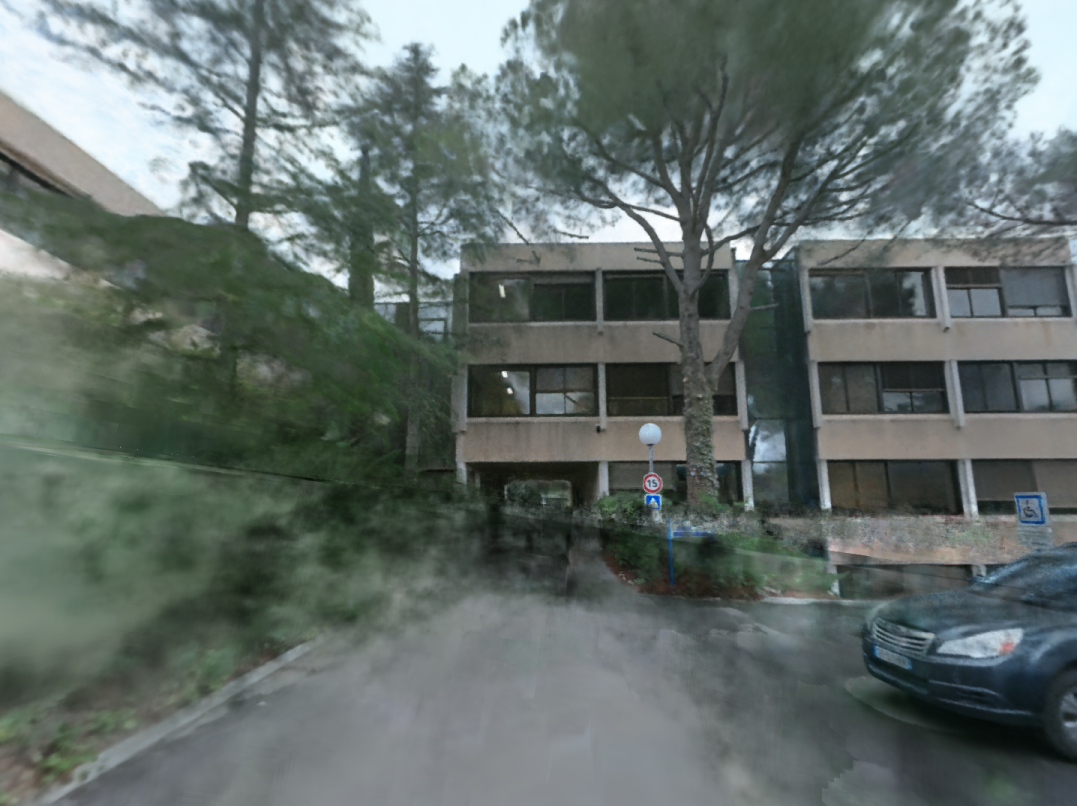} 
	\end{tabular}
	\caption{
		\label{fig:compare}
		Qualitative comparison of our method to previous solutions on a single chunk. We show a reasonably reduced level-of-detail ($\tau_\epsilon = 6$ pixels) for ours.
	}
\end{figure*}

\section{Results and Evaluation}
\label{sec:eval}

We demonstrate our method on our four captured scenes: \textsc{SmallCity}, \textsc{BigCity}, \textsc{Campus} 
and also on one scene provided by Wayve (see Fig.~\ref{fig:chunks} and Fig.~\ref{fig:compare}). 
\subsection{Results}

The results are best appreciated in the supplemental videos, where we see that we can navigate in the captured large scenes in real-time.
The paths we show in the video illustrate the full extent of captured area to show how large they are. For each scene we stop and show a free-viewpoint walk-around session. This works particularly well in areas where camera density is higher, such as places where multiple passes of capture crossed, etc.

Our smallest scene, \textsc{SmallCity}, contains only a single urban block. We cover more than a kilometer distance in the \textsc{Campus, Wayve} scenes, and several kilometers in the \textsc{BigCity} scene. During training, they are subdivided into 4, 11, 22, and 46 chunks, respectively. Each chunk has 2--8M leaf Gaussians. The coarse, single chunk, and hierarchy optimization take one hour each on one NVIDIA V100 GPU (32\,GB), with 2--5\,GB, 8--10\,GB, and 11--16\,GB peak memory usage respectively. After per-chunk optimization and consolidation, disk sizes for the hierarchical scenes are 6, 17, 27, and 88\,GB, $\approx$68\% larger than non-hierarchical 3DGS files.
While the quality is globally good, there are some artifacts.
Most such visual artifacts are due to the sparse nature of these large captures: contrary to traditional radiance field captures, a given point in the scene is only seen by a handful of cameras. Other artifacts
are due to distractors in the scene that were not completely removed: we are able to remove most moving cars, but cars coming to a stop are not correctly identified, leading to inconsistent data and bad optimization. Developing a complete solution to this problem is orthogonal to our contribution (see also~Sec.~\ref{sec:discuss} and Appendix \ref{sec:data}, \ref{sec:pose}).

We also show the effect of progressive interpolation between hierarchy levels in the video.

\subsection{Evaluation}

Comparing to other methods is difficult, since no other approach with code available can handle our large scenes.
We thus decided to compare the quality of our method, given different options on a single chunk that can be reasonably handled by previous methods. All methods, including ours, receive the same input, i.e., only the images that belong to that chunk.
In particular, we compare to F2-NeRF, Instant-NGP, the original 3DGS, Mip-NeRF 360 and Mega-NeRF~\cite{Turki_2022_meganerf} on one chunk of each dataset.
Finally, we perform ablations on several elements in our method, illustrating the importance of each corresponding algorithmic component.
All real-time rendering results and performance metrics were collected on an NVIDIA A6000 GPU.

\mparagraph{Comparisons to Other Methods.}
In Fig.~\ref{fig:compare}, and supplemental videos, we show our method compared to previous approaches for a single chunk, illustrating the visual results of the quantitative comparisom below. These show the benefit of our two main improvements over 3DGS for such sparse capture, namely depth supervision and the modified densification.

\begin{table*}
\caption{
\label{tab:comparisons}
	We show quality and frames per second (FPS) for rendering with our method compared to previous work, for one chunk per scene, since competitors cannot directly handle our full scenes. "Ours (leaves)" is the output of  our per-chunk optimization, including the improvements in ~Sec.~\ref{sec:chunk-train}, running on a single chunk with the same input data as previous solutions.
	We further evaluate the effect of our novel level-of-detail components (see Sec.~\ref{sec:methodlod}) on quality. "Ours ($\tau$)" indicates our method using the generated hierarchy for each chunk, for different granularity settings ($\tau_1 = 3$, $\tau_2 = 6$ and $\tau_3 = 15$ pixels, respectively). "Ours opt" is the same, but after the optimization of the hierarchy (Sec.~\ref{sec:post-optim}), again with different granularity settings. We highlight \textbf{best} and \underline{second-best} in each category.}
\setlength{\tabcolsep}{4pt}
\scalebox{0.95}
{\begin{tabular}{l|cccc|cccc|cccc|cccc}
 \multicolumn{1}{r|}{Scene} & \multicolumn{4}{c|}{\textsc{SmallCity}} & \multicolumn{4}{c|}{\textsc{Wayve}} & \multicolumn{4}{c|}{\textsc{Campus}} & \multicolumn{4}{c}{\textsc{BigCity}} \\
	Method & \small{PSNR$^\uparrow$} & \small{LPIPS$^\downarrow$}  & \small{SSIM$^\uparrow$} & FPS$^\uparrow$ & \small{PSNR$^\uparrow$} & \small{LPIPS$^\downarrow$}  &  \small{SSIM$^\uparrow$} & FPS$^\uparrow$ & \small{PSNR$^\uparrow$}  & \small{LPIPS$^\downarrow$}  &  \small{SSIM$^\uparrow$} & FPS$^\uparrow$ & \small{PSNR$^\uparrow$} & \small{LPIPS$^\downarrow$} & \small{SSIM$^\uparrow$} & FPS$^\uparrow$ \\
\hline 
Mip-NeRF 360 & 24.70 & 0.348 & 0.765 & - & 25.05 & 0.328 & 0.767 & - & 20.95 & 0.442 & 0.731 & - & 20.95 & 0.442 & 0.731 & - \\
INGP-big   & 23.47 & 0.426 & 0.715 & - & 22.84 & 0.382 & 0.711 &  - & 20.37 & 0.476 & 0.700 & - & 19.75 & 0.516 & 0.674 & - \\ 
F2-NeRF-big & 24.53 & 0.342 & 0.762 & - &24.10 & 0.320 & 0.758 &  - & 19.46 & 0.449 & 0.710 & - & 19.25 & 0.478 & 0.681 & -  \\
3DGS       & 25.34 & 0.337 & 0.776 & 99 &25.13 & 0.299 & 0.797 & 125 & 23.87 & 0.378 & 0.785 & 82 & 21.48 & 0.445 & 0.721 & 72  \\
Ours (leaves)      & \textbf{26.62} & \textbf{0.259} & \textbf{0.820} & 58 &\textbf{25.35} & \textbf{0.256} & \textbf{0.813} & 70 & \textbf{24.61} & \textbf{0.331} & \textbf{0.807} & 51 & \textbf{23.10} & \textbf{0.348} & \textbf{0.769} & 39  \\
Ours ($\tau_1$) 	& 26.49 & 0.264 & 0.817 & 87 & \underline{25.26} & 0.258 & 0.811 & 90 & 24.58 & 0.334 & 0.805 & 68 & 23.09 & 0.350 & 0.768 & 76  \\
Ours opt ($\tau_1$) & \underline{26.53} & \underline{0.263} & \underline{0.817} & 86 & 25.25 & \underline{0.258} & \underline{0.811} & 84 & \underline{24.59} & \underline{0.333} & \underline{0.806} & 64 & \underline{23.09} & \underline{0.350} & \underline{0.768} & 78  \\
Ours ($\tau_2$) 	& 25.72 & 0.297 & 0.796 & 106 & 24.63 & 0.279 & 0.792 & 107 & 24.33 & 0.352 & 0.792 & 79 & 22.97 & 0.365 & 0.758 & 101  \\
Ours opt ($\tau_2$) & 26.29 & 0.275 & 0.810 & 110 & 25.03 & 0.270 & 0.803 & 110 & 24.50 & 0.340 & 0.801 & 80 & 23.05 & 0.359 & 0.762 & 102  \\
Ours ($\tau_3$) 	& 23.04 & 0.423 & 0.699 & \underline{157} & 22.40 & 0.359 & 0.714 & \underline{125} & 22.93 & 0.427 & 0.736 & \textbf{106} & 22.19 & 0.437 & 0.710 & \underline{128}  \\
Ours opt ($\tau_3$) & 25.68 & 0.324 & 0.786 & \textbf{159} & 24.49 & 0.308 & 0.775 & \textbf{135} & 24.12 & 0.378 & 0.780 & \underline{104} & 22.82 & 0.402 & 0.742 & \textbf{137}  \\
\end{tabular}}
\end{table*}

\begin{table}
		\caption{\label{tab:meganerf}
			Comparisons on Mega-NeRF's Mill 19 aerial dataset.}
		\setlength{\tabcolsep}{4pt}
		{\begin{tabular}{l|ccc|ccc}
				\multicolumn{1}{r|}{Scene} & \multicolumn{3}{c|}{\textsc{Building}} & \multicolumn{3}{c}{\textsc{Rubble}} \\
				Method & \small{PSNR$^\uparrow$} & \small{LPIPS$^\downarrow$}  & \small{SSIM$^\uparrow$} & \small{PSNR$^\uparrow$} & \small{LPIPS$^\downarrow$}  &  \small{SSIM$^\uparrow$}  \\
				\hline 
				Mega-NeRF & 20.93 & 0.504 & 0.547 & 24.06 & 0.516 & 0.553 \\
				Ours opt ($\tau_2$) & \textbf{21.52} & \textbf{0.297} & \textbf{0.723} & \textbf{24.64} & \textbf{0.284} & \textbf{0.755}  
		\end{tabular}}
\end{table}

\begin{table*}
		\caption{3DGS render time (ms) breakdown for our single-chunk scenes with and without our hierarchy. Compared to 3DGS, our LOD mechanism includes two additional stages: \texttt{cut/expand*} and \texttt{weights}. Note that \texttt{cut/expand*} need not be called each frame and runs asynchronously to rendering.}
		\scalebox{0.94}{
			\begin{tabular}{l|ccc|ccc|ccc|ccc}	
				\multicolumn{1}{r|}{Scene}& \multicolumn{3}{c|}{\textsc{SmallCity}} & \multicolumn{3}{c|}{\textsc{Wayve}} & \multicolumn{3}{c|}{\textsc{Campus}} & \multicolumn{3}{c}{\textsc{BigCity}}\\
				Stage & 3DGS & Ours (leaves) & Ours ($\tau_2$) & 3DGS & Ours (l.) & Ours ($\tau_2$) & 3DGS & Ours (l.) & Ours ($\tau_2$) & 3DGS & Ours (l.) & Ours ($\tau_2$)\\
				\hline
				\texttt{cut/expand*} &
				- & 20.34 & 7.67 & - &12.13 & 7.65 & - & 17.23 & 10.48 & - & 24.19 & 8.25\\
				\hline
				\texttt{weights} &
				- & 2.40 & 3.52 & - & 2.46 & 3.46 & - & 2.44 & 3.05 & - & 1.76 & 3.55\\
				\texttt{preprocess} & 1.28 & 3.26 & 1.34 & 0.90 & 2.38 & 1.07
				& 1.17 & 3.26 & 1.67 & 1.32 & 4.16 & 1.21\\
				\texttt{duplicate} & 0.54 & 0.58 & 0.58 & 0.85 & 0.83 & 0.95
				& 1.12 & 1.21 & 1.24 & 0.84 & 0.93 & 0.90\\
				\texttt{tile ranges} & 0.09 & 0.09 & 0.07 & 0.10 &0.10 & 0.99 
				& 0.15 & 0.15 & 0.14 & 0.11 & 0.12 & 0.10\\
				\texttt{alpha-blend} & 8.96 & 13.20 & 3.41 & 2.73 & 4.66 & 3.00
				& 5.17 & 8.46 & 3.98 & 7.58 & 15.23 & 3.33\\
		\end{tabular}}
		\label{tab:runtime}
\end{table*}

\begin{table*}[!h]
		\caption{Resource and performance analysis using our full-scene camera paths. For different granularities $\tau_\epsilon$, we report per-frame averages for Gaussians rendered (i.e., Gaussians required on-chip) \emph{\#Render}, requested Gaussians transferred from CPU to GPU \emph{\#Trans.}, and achieved frames per second \emph{FPS}. To illustrate the benefits of our method, we also report rendered Gaussians as a percentage of all leaves, i.e., the number of Gaussians that 3DGS would render.}
		\begin{tabular}{l|ccc|ccc|ccc|ccc}	
			\multicolumn{1}{r|}{Scene}& \multicolumn{3}{c|}{\textsc{SmallCity}} & \multicolumn{3}{c|}{\textsc{Wayve}} & \multicolumn{3}{c|}{\textsc{Campus}} & \multicolumn{3}{c}{\textsc{BigCity}}\\
			$\tau_\epsilon$ & \#Render (\%) & \#Trans & FPS & \#Render (\%) & \#Trans & FPS & \#Render (\%) & \#Trans & FPS & \#Render (\%) & \#Trans & FPS\\
			\hline
			$\tau_1 = 3$ px & 9.44M (66\%) & 1586 & 46 & 12.4M (34\%) & 2170 & 40 & 21.6M (33\%) & 9314 & 32 & 17.6M (19\%) & 5750 & 31\\
			$\tau_2 = 6$ px & 5.64M (39\%) & 1795 & 78 & 6.39M (17\%) & 2208 & 65 & 10.2M (16\%) & 7996 & 62 & 8.21M (8\%) & 5355 & 56\\
			$\tau_3 = 15$ px& 2.26M (16\%) & 1536 & 150 & 2.10M (6\%) & 1614 & 125 & 3.01M (5\%) & 5147 & 133 & 2.68M (3\%) & 4167 & 103\\
		\end{tabular}
		\label{tab:numbers}
\end{table*}

We perform quantitative evaluation on our datasets by excluding from training every 50th image alphabetically from each camera in the rig, which are then used for testing. 
We compute standard error metrics PSNR, LPIPS and SSIM for each method on the single chunk (see Tab.~\ref{tab:comparisons}). The first part on the table shows results for our method running only on one chunk, without the hierarchy, given only the cameras of the chunk. We disable exposure optimization for this experiment as other approaches do not take exposure change into account by default. 
This is a ``handicapped'' comparison for us, since the strength of our method is that it can handle the entire scene, however it allows a ``fair'' comparison to others since every method starts with the same data. We see that our method outperforms all previous methods for chunks that are in most cases larger than those used in previous methods.

To assess the flexibility of our method, we evaluate the result of our single chunk optimization on the established small-scale Mip-NeRF 360 dataset. We note that our method \emph{is not} targeted at such scenes.
In comparison to original 3DGS, our single-chunk optimization achieves similar quality on average over the entire dataset, yielding PSNR of 29.11/28.96\,db for theirs/ours. The difference to 3DGS is due to tuning the training for chunk-sized scenes. For LOD levels $\tau_1, \tau_2$ and $\tau_3$, PSNR results with basic/optimized hierarchies are 28.86/28.87\,db, 28.05/28.52\,db, and 24.82/27.25\,db, respectively.

Given that no other method can treat street-level data at this scale, as a best effort we
compare to Mega-NeRF~\cite{Turki_2022_meganerf}, which handles extensive scenes, albeit tested on aerial photography. We compare our method to theirs on the authors' proposed Mill 19 dataset.
We use pixSFM~\cite{lindenberger2021pixsfm} camera poses provided by the Mega-NeRF data release.
Since these datasets lack SfM points, we use the COLMAP matcher (with 100 neighbor frames) and triangulator to generate 3D points given the provided poses. We then scale the scenes to metric units. 
We split the scene using 200$\times$200\,m chunks, leading to 2 and 4 chunks for the \textsc{Building} and \textsc{Rubble} scenes, respectively.
We downsample the images four times and optimize the exposure affine transforms for test views by including the left half of the test images in the training set, using the remaining half for testing, all in accordance with Mega-NeRF's code release. Tab.~\ref{tab:meganerf} shows that our method compares favorably, despite not being tuned for aerial data. 
Most importantly, Mega-NeRF reports training times from 27 to 30 hours on eight NVIDIA V100 GPUs, while our method takes 3 hours on two (\textsc{Building}) and four (\textsc{Rubble}) V100s and achieves real-time rendering.

\mparagraph{Evaluating the Quality of the Hierarchy.}
In the second part of Tab.~\ref{tab:comparisons}, we evaluate the various options of our method by running our hierarchical method on the full scene, and providing quantitative results for the same chunk. In particular, we show the effect on image metrics when rendering with the unoptimized hierarchy ("Ours") for different target granularities.
As expected, when moving higher up in the hierarchy, quality drops. For the optimized hierarchy (Sec.~\ref{sec:post-optim}, "Ours opt"), we see that at the finest granularity target, the solutions have similar performance. However, when choosing a coarser cut in the hierarchy, the optimization improves the result, which was the main goal of this step. In practice, this means that for a given computational budget, optimized hierarchies improve visual quality.

\mparagraph{Performance Analysis.}
To analyze our rendering performance in the single-chunk scenes, Tab. \ref{tab:runtime} provides detailed breakdowns of the time spent in different stages with the original 3DGS rendering and our LOD-enabled prototype. In addition to the original pipeline, we compute the interpolation weights in each frame (\texttt{weights}), as described in Sec. \ref{sec:select}. This incurs an additional cost in the range from 1.5 to 4\,ms. We observe an overhead for Ours (leaves) compared to 3DGS, both in the \texttt{preprocess} (i.e., projection and evaluation of splat properties) and the \texttt{alpha-blend} stage. The former is due to loading twice as much data to produce interpolated Gaussian attributes. The latter is caused by the computation of auxiliary blending weight $\alpha'$ with a comparatively expensive $\texttt{pow}$ instruction. For $\tau_\epsilon = 6$, however, our LOD mechanism results in a significantly reduced workload, thus we can \emph{accelerate} these stages compared to 3DGS. \texttt{cut/expand*} simultaneously updates the cut and enqueues (future) required, higher-detail Gaussians for transfer. Although it is comparably slow, in practice, this stage runs asynchronously to rendering, and therefore does not impact real-time performance.

\mparagraph{Runtime Analysis for Large Scene Rendering.}
We have evaluated speed and resource use of our method when displaying the full, large-scale scenes, following the camera paths shown in our accompanying video. All paths yielded 30+ FPS on average for the high-quality setting, $\tau_1$, and $\approx$60 FPS at our medium setting, $\tau_2$. We note that due to their size, neither \textsc{Campus} nor \textsc{BigCity} would run with the original 3DGS renderer on our test system; \textsc{Wayve} exceeds the capacities of a NVIDIA RTX 4090 and \textsc{SmallCity} those of a mid-range NVIDIA RTX 4080. Tab.~\ref{tab:numbers} reports the number of rendered Gaussians on the recorded paths as total count and percentage of what 3DGS would (theoretically) need to process.  Our LOD mechanism effectively curbs load and memory consumption; the larger the scene, the higher the reduction. Node and bound information raises the theoretical per-Gaussian memory footprint to 284 bytes. Our current implementation, including convenience structs, uses 400 bytes (69\% more than 3DGS) per Gaussian. We also report the average number of Gaussians transferred per frame.

\newcommand{\spyimg}[4]{%
	\begin{tikzpicture}[spy using outlines={circle,red,magnification=4,size=1.4cm, connect spies}]
		\node[anchor=south west,inner sep=0] at (0,0) {\includegraphics[height=#1]{#2}};
		\spy on (#3) in node [left] at (#4);
	\end{tikzpicture}%
}

\begin{figure*}[!h]
	\setlength{\tabcolsep}{1pt}
	\begin{tabular}{cccccc}
		& Consolidation & Depth Regularization & Chunk Bundle Adj. & Exposure Comp. & Hierarchy Optimization \\
		\raisebox{6mm}{	\rotatebox{90}{Without}} & \includegraphics[height=2.52cm]{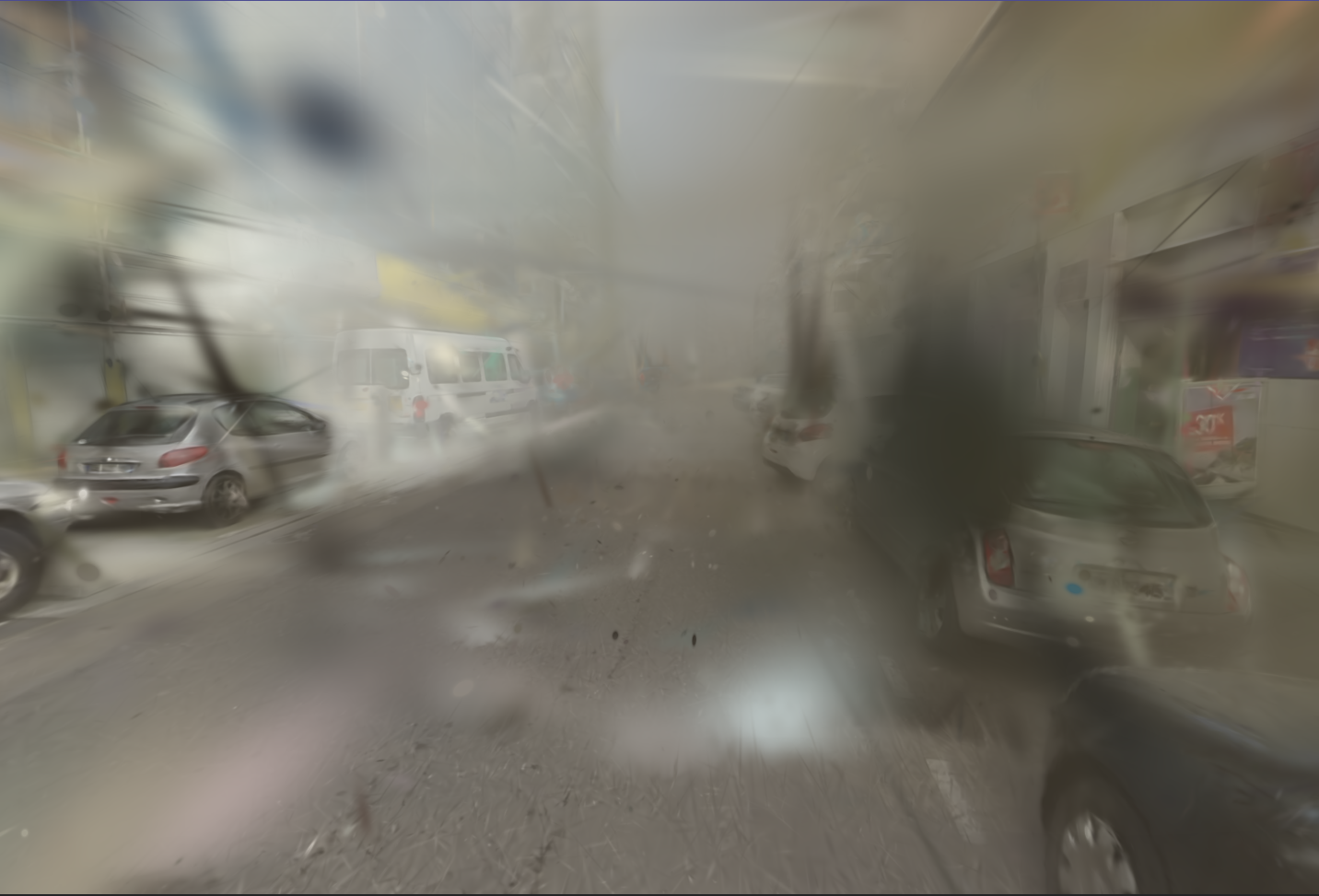} & \includegraphics[height=2.52cm]{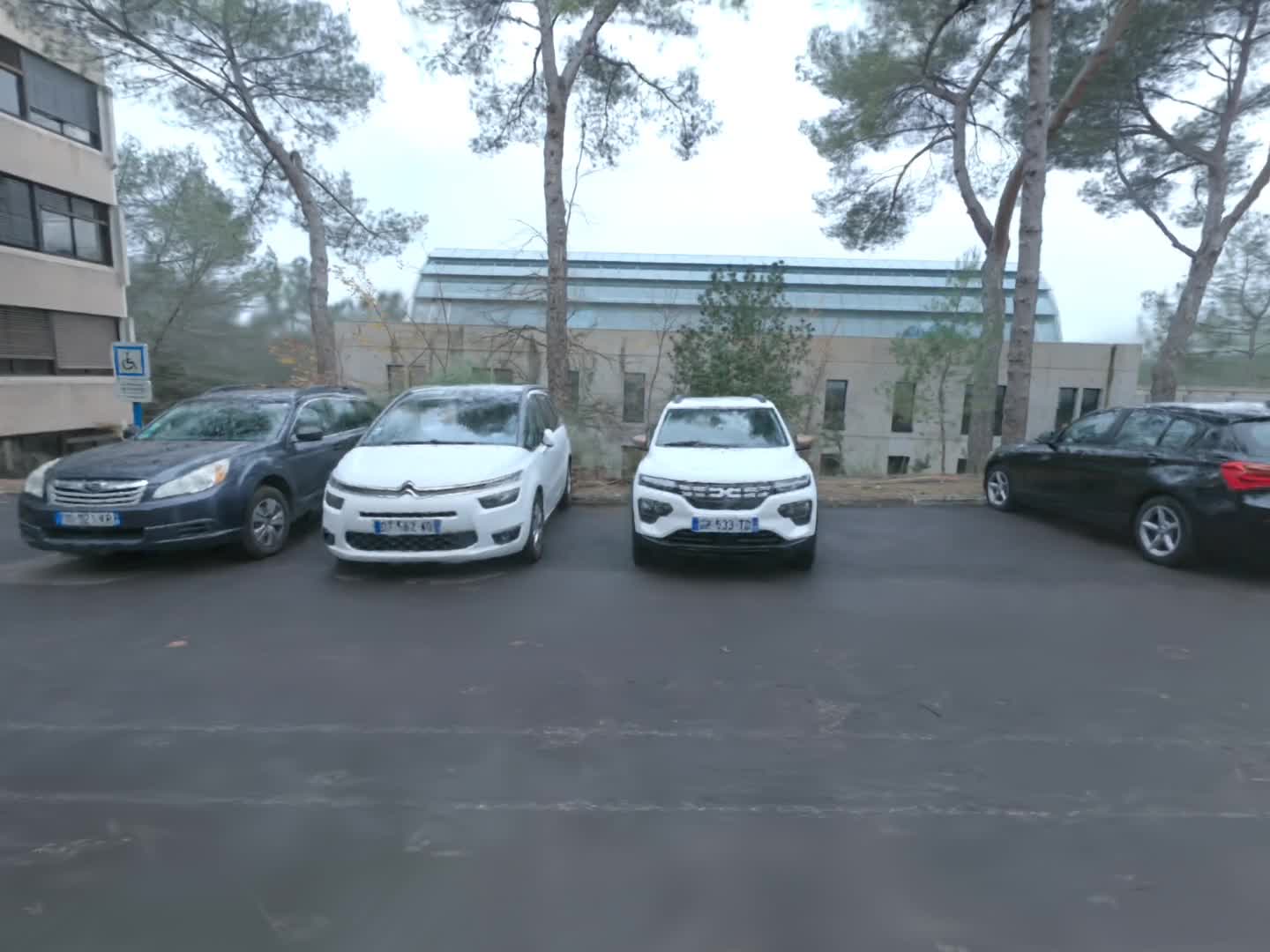} & \includegraphics[height=2.52cm]{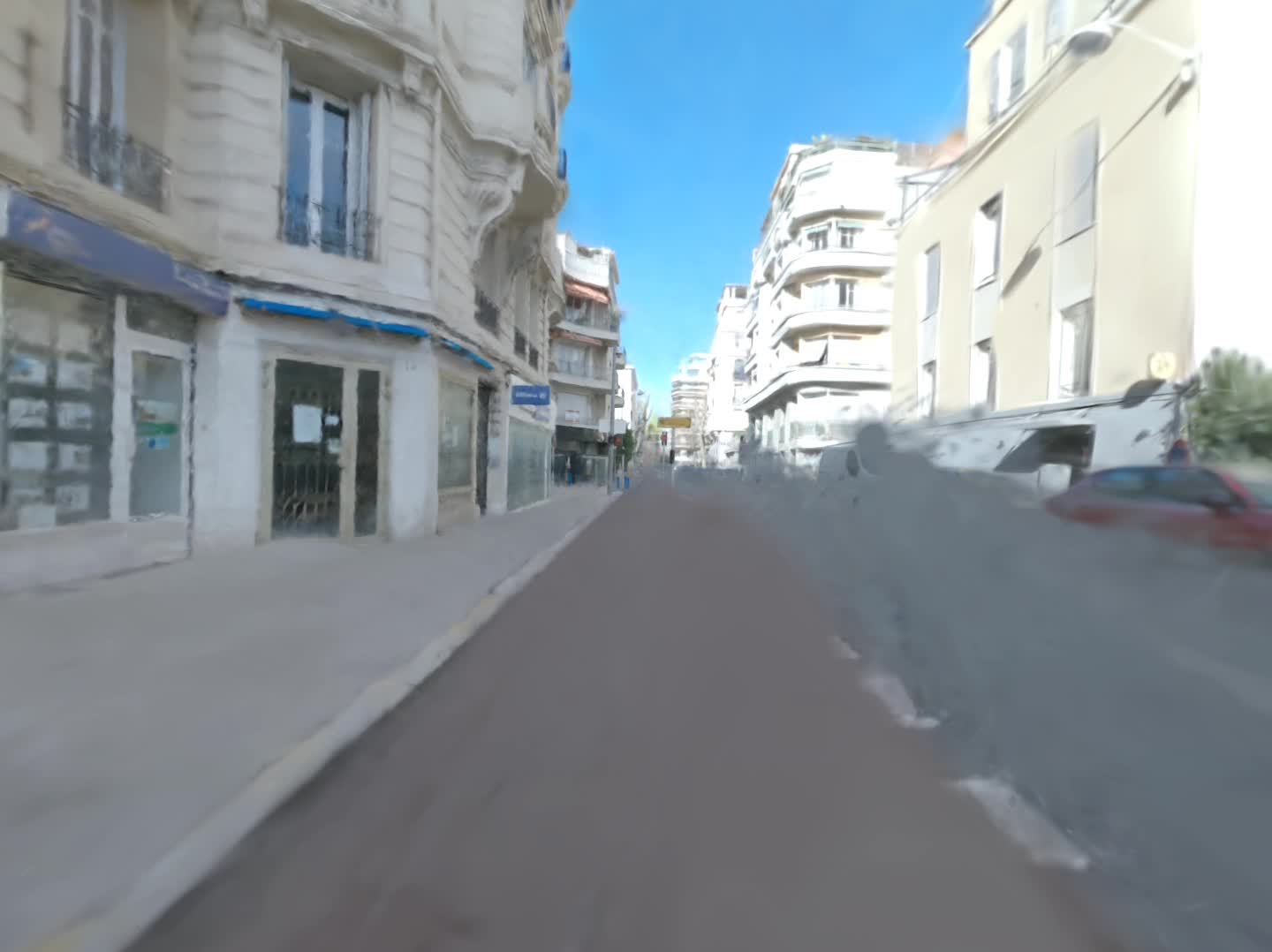} &
		\includegraphics[height=2.52cm]{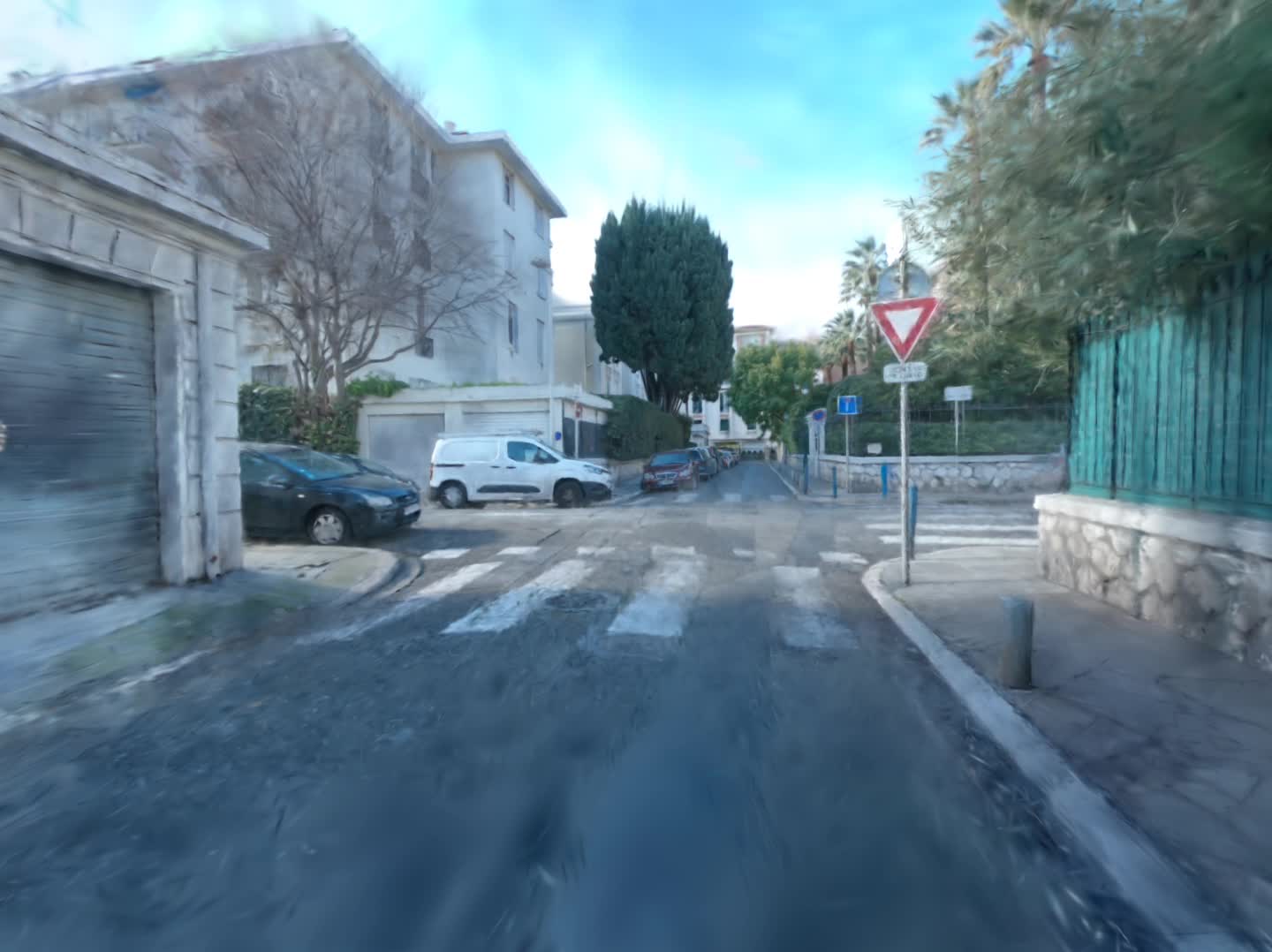} & 				\spyimg{2.52cm}{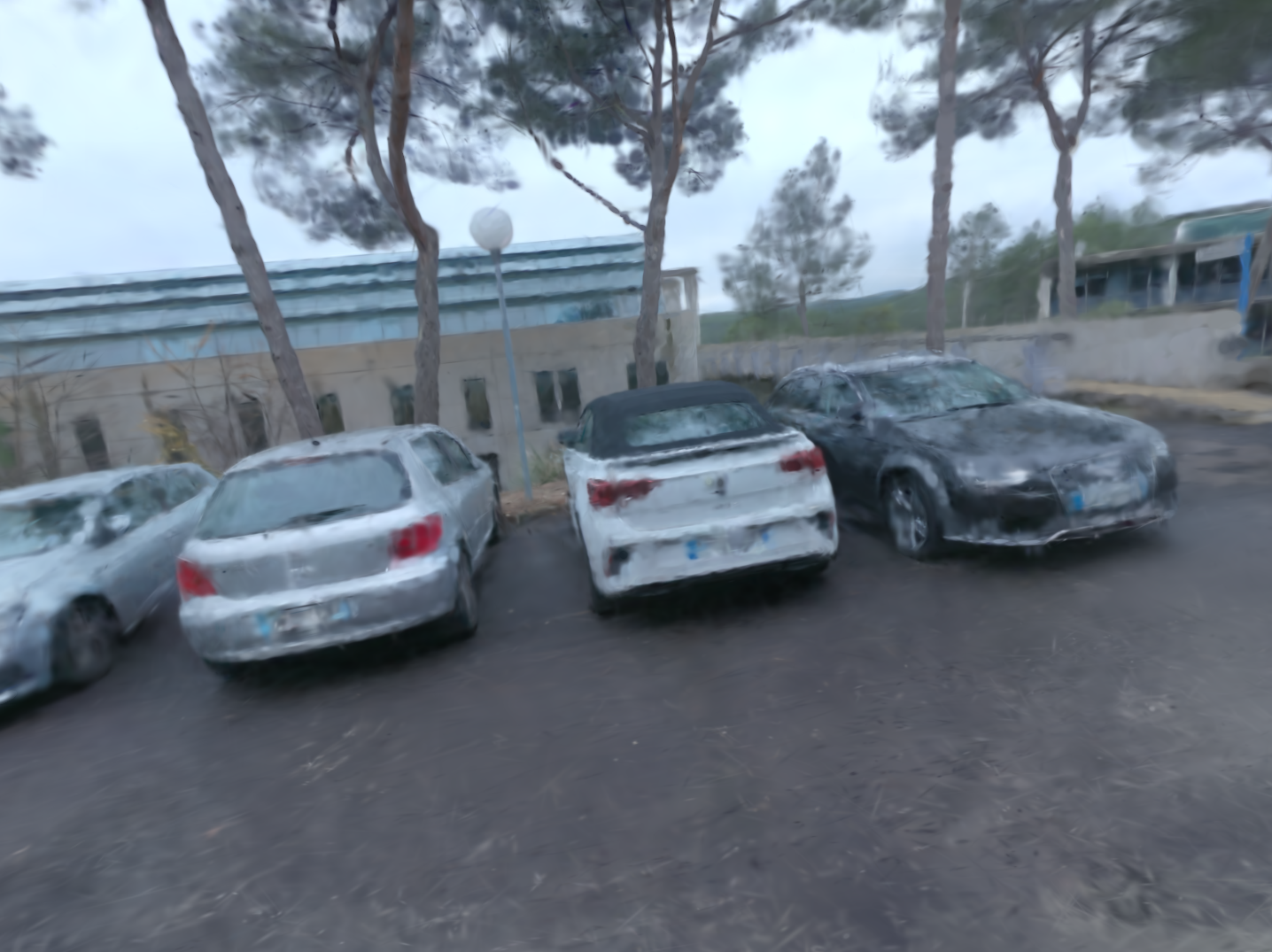}{1.65,1.3}{1.4,2.0}\\
		\raisebox{9mm}{	\rotatebox{90}{With}} & \includegraphics[height=2.52cm]{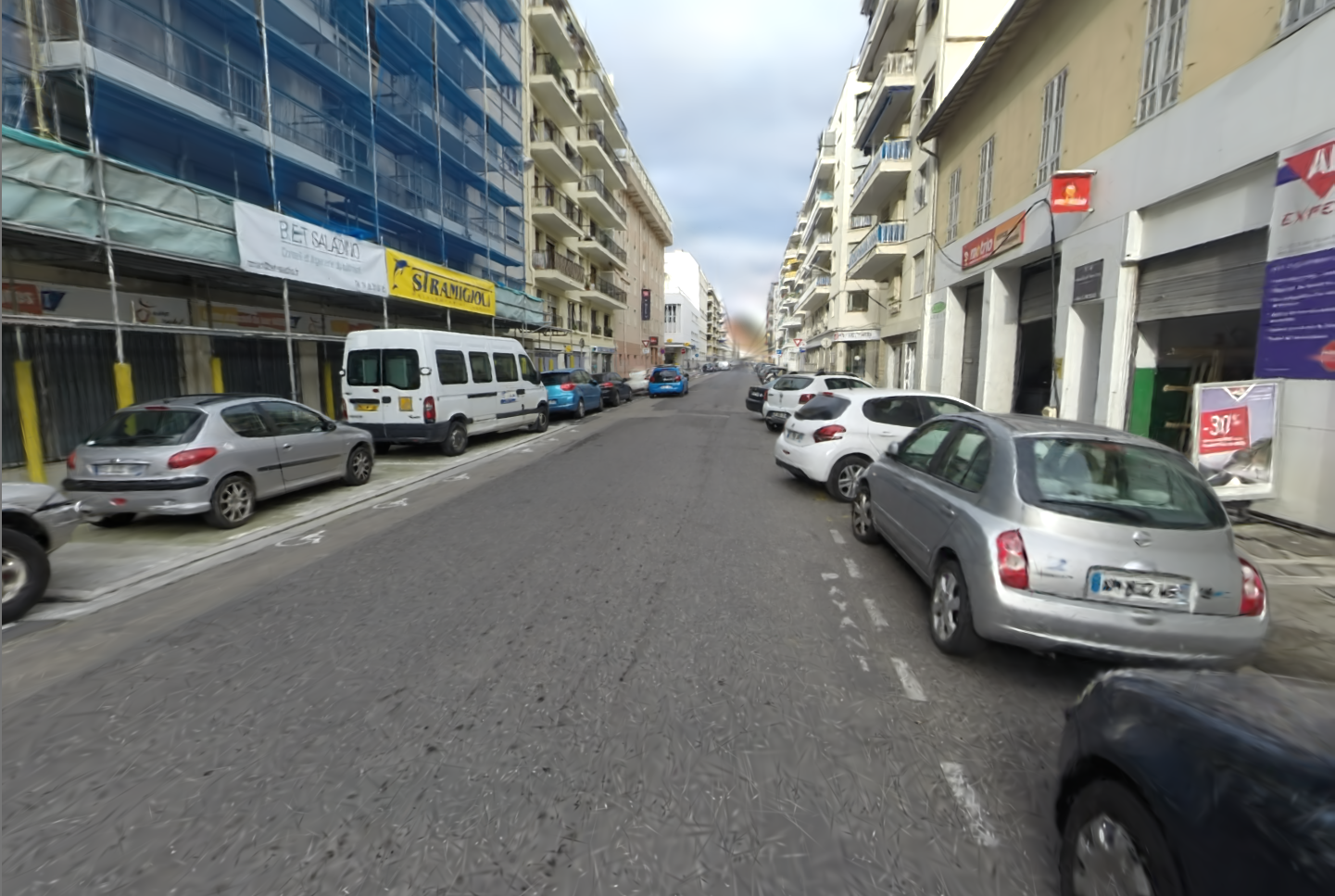} &
		\includegraphics[height=2.52cm]{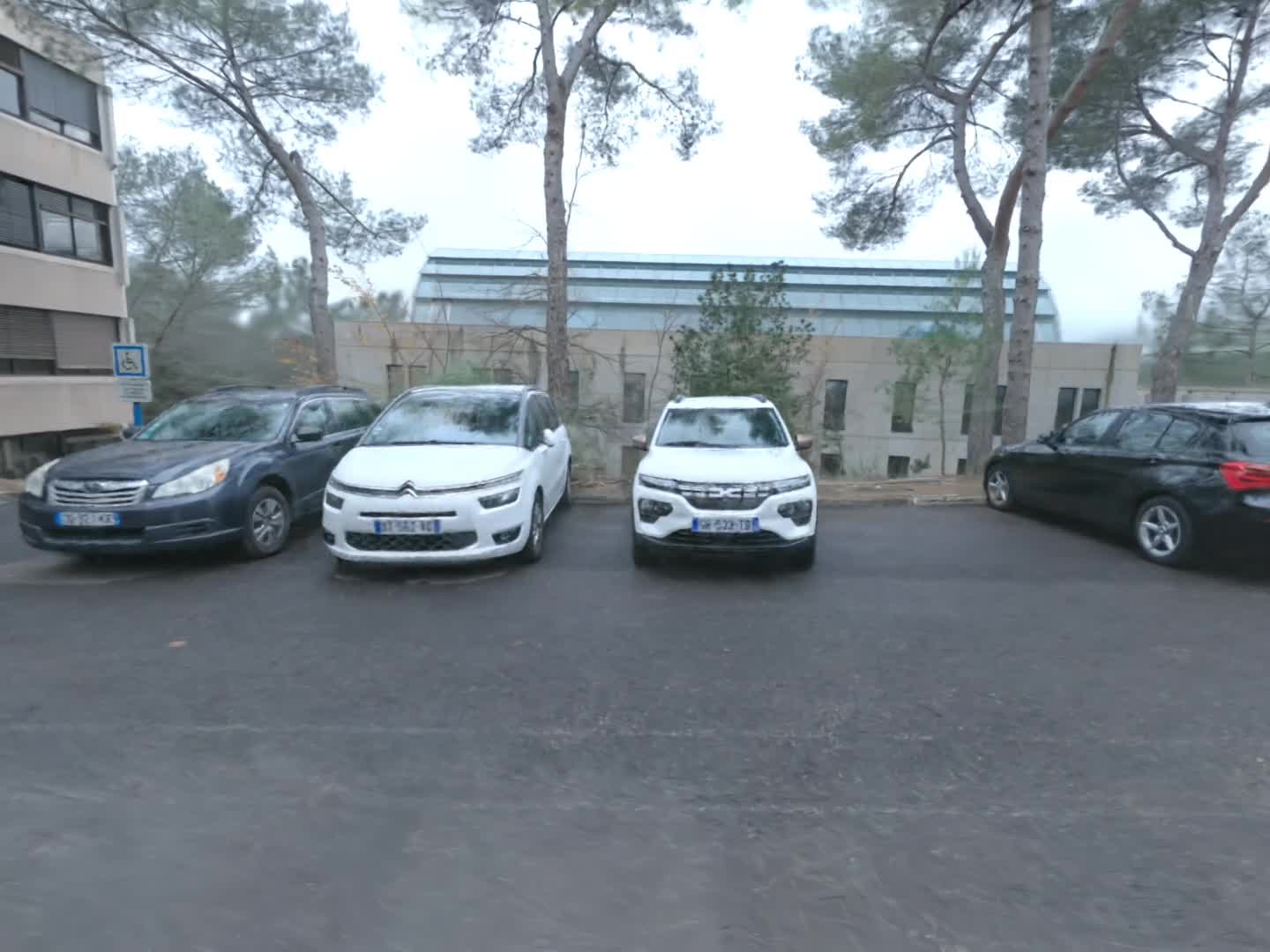}   &
		\includegraphics[height=2.52cm]{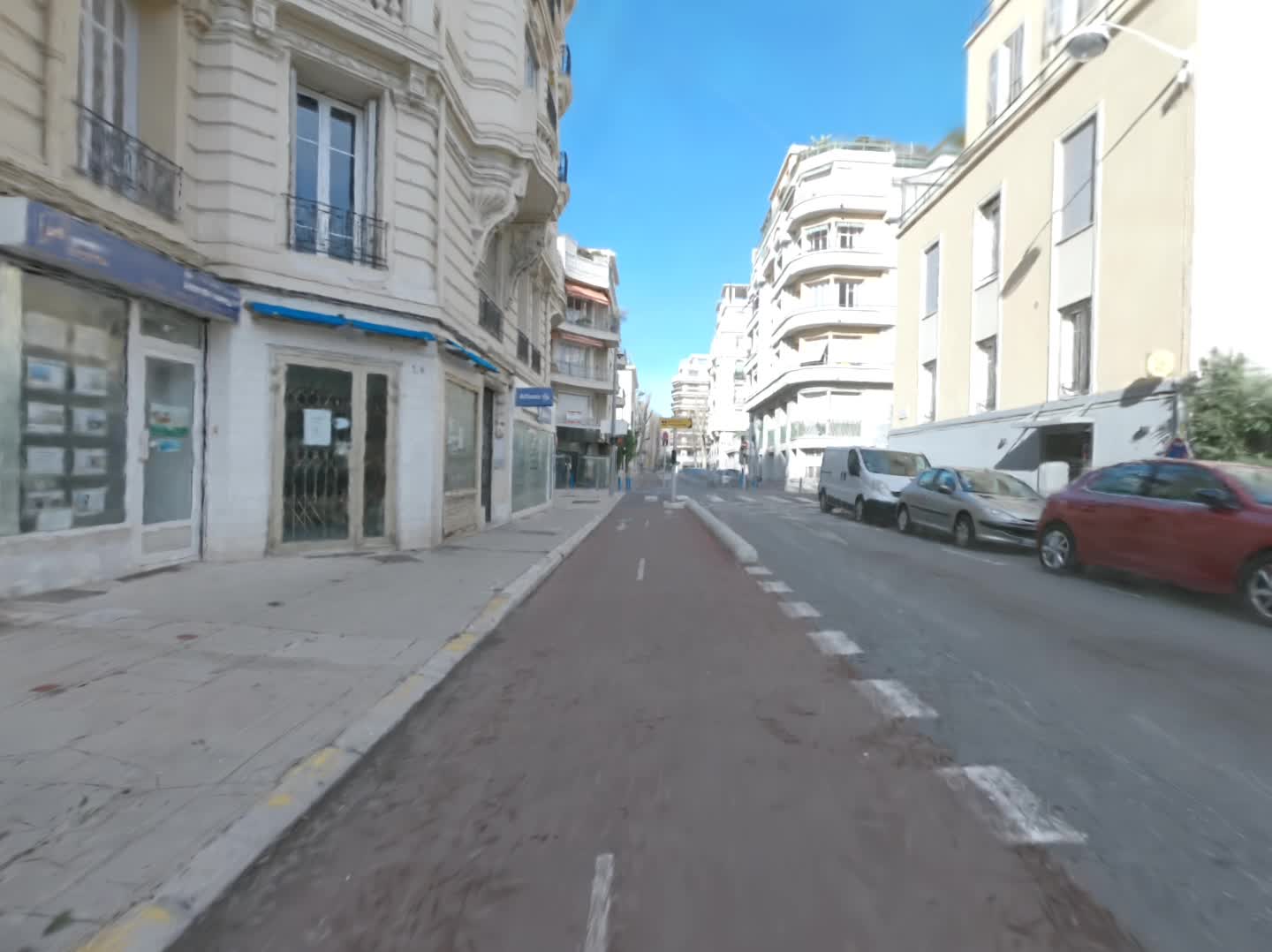} & 
		\includegraphics[height=2.52cm]{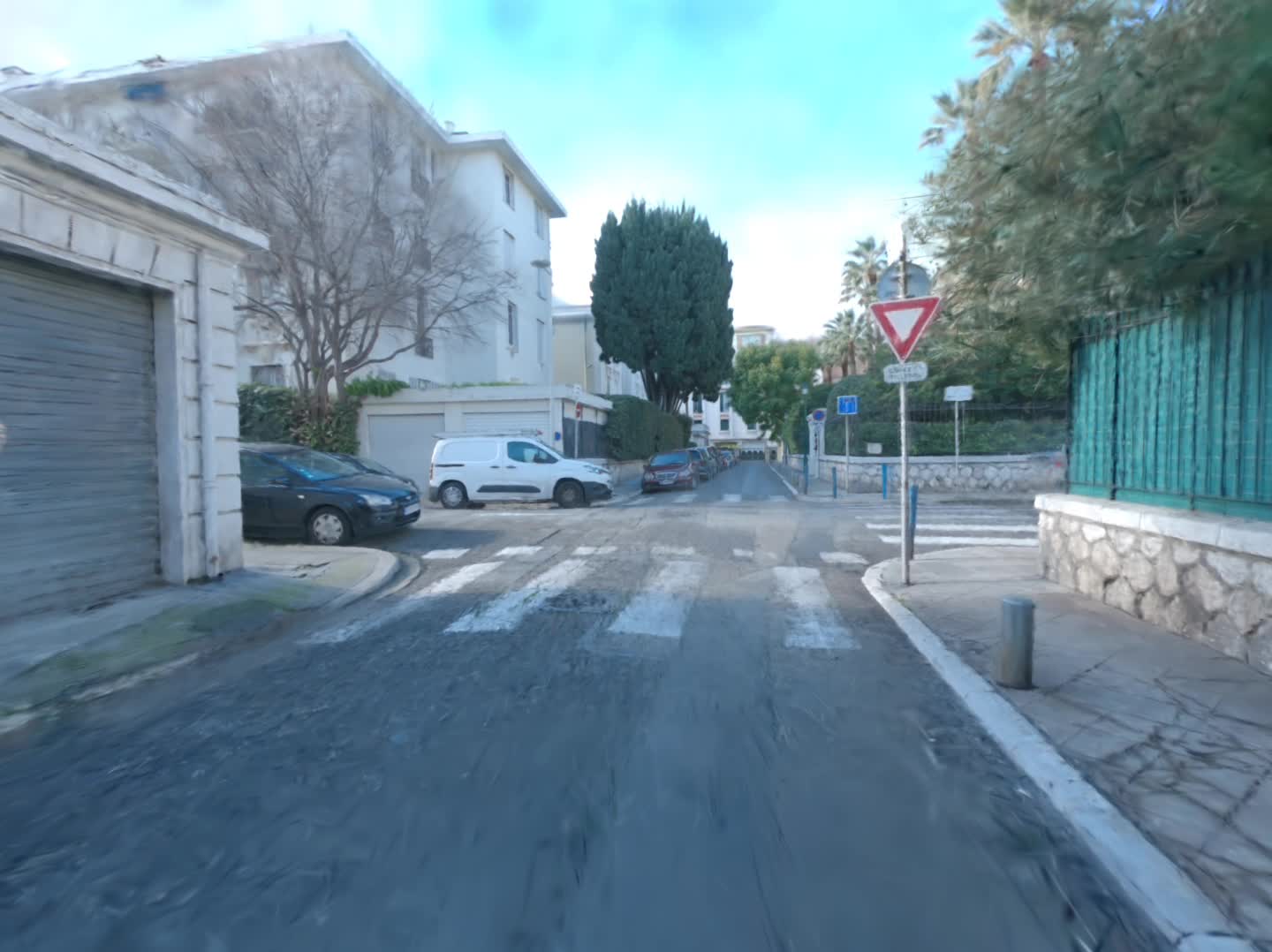} & 
		\spyimg{2.52cm}{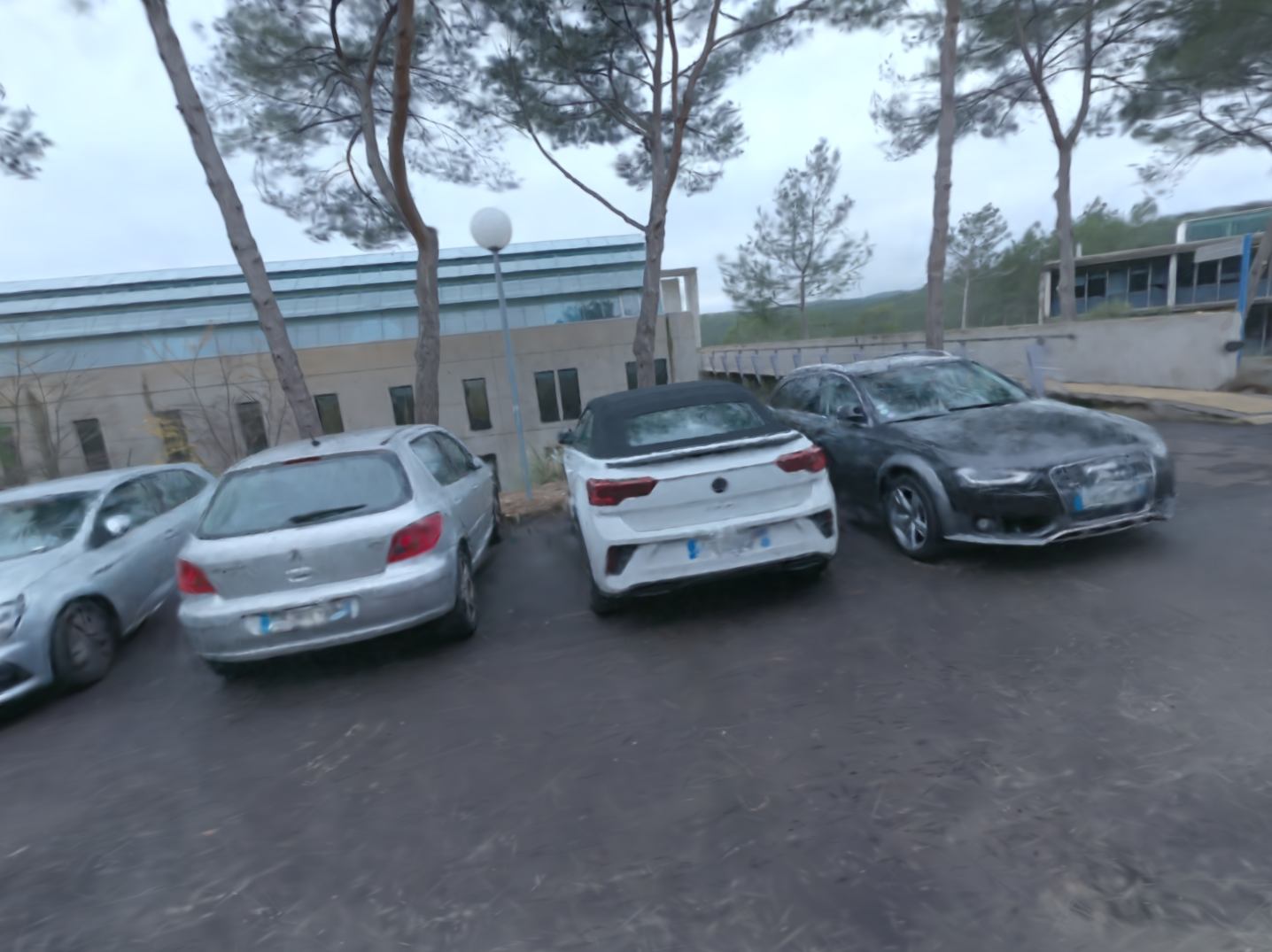}{1.65,1.3}{1.4,2.0}
	\end{tabular}
	\caption{
		\label{fig:ablation}Qualitative ablations.
		Column 1. Top: Without removing unwanted Gaussians in consolidation; Bottom: With unwanted Gaussians removed. We can clearly see the improvement in quality.
		Column 2. Top: Result without depth supervision. Bottom: result on the same view with depth supervision, which improves densification, particularly on the road.
		Column 3. Top: Result using COLMAP hierarchical mapper; Bottom: After our additional per-chunk bundle adjustment step that improves the results significantly.
		Column 4. Top: Without exposure handling; Bottom: With per-image exposure optimization.
		Column 5. Top: Without optimization after hierarchy generation, rendered with $\tau_\epsilon = 15$ pixels; Bottom: With optimization.
	}
\end{figure*}

\begin{table*}[!h]
		\caption{
			\label{tab:ablation}
			Quantitative ablations. Using 3DGS as a baseline, we assess how depth supervision and our modified, maximum-based densification impact quality. We highlight the \textbf{best} and \underline{second-best} results in each category.
		}
		\scalebox{0.879}{
			\begin{tabular}{l|ccc|ccc|ccc|ccc|ccc}
				\multicolumn{1}{r|}{Scene} &
				\multicolumn{3}{c|}{\textsc{SmallCity}} & \multicolumn{3}{c|}{\textsc{Wayve}} &  \multicolumn{3}{c|}{\textsc{Campus}} & \multicolumn{3}{c|}{\textsc{BigCity}} & \multicolumn{3}{c}{\textsc{Average}} \\
				Method & \small{PSNR$^\uparrow$} & \small{LPIPS$^\downarrow$}  & \small{SSIM$^\uparrow$} & \small{PSNR$^\uparrow$} & \small{LPIPS$^\downarrow$}  & \small{SSIM$^\uparrow$} & \small{PSNR$^\uparrow$} & \small{LPIPS$^\downarrow$}  & \small{SSIM$^\uparrow$} & \small{PSNR$^\uparrow$} & \small{LPIPS$^\downarrow$}  & \small{SSIM$^\uparrow$} & \small{PSNR$^\uparrow$} & \small{LPIPS$^\downarrow$}  & \small{SSIM$^\uparrow$} \\
				\hline
				3DGS							& 25.34 & 0.337 & 0.776 & 25.13 & 0.299 & 0.797  & 23.87 & 0.378 & 0.785 & 21.48 & 0.445 & 0.721 & 23.96 & 0.365 & 0.770 \\
				Ours \small{w/o modif. dens.} 	& 25.54 &	0.350 &	0.783 & 24.93 & 0.295 &	0.773 &	24.14 &	0.410 &	0.766 & 23.10 & 0.429 &	0.740 &	24.43 &	0.371 &	0.766 \\
				Ours \small{w/o depth reg.} 	& \textbf{26.66} &	\underline{0.261} &	\underline{0.818} & \textbf{25.86} & \textbf{0.245} &	\textbf{0.825} &	\textbf{24.74} &	\underline{0.337} &	\underline{0.805} & \textbf{23.15} & \underline{0.350} &	\underline{0.769} &	\textbf{25.10} &	\textbf{0.298} &	\textbf{0.804} \\
				Ours 							& \underline{26.62} & \textbf{0.259} & \textbf{0.820} & \underline{25.35} & \underline{0.256} & \underline{0.813} & \underline{24.61} & \textbf{0.331} & \textbf{0.807} & \underline{23.10} & \textbf{0.348} & \textbf{0.769} & \underline{24.98} & \underline{0.303} & \underline{0.798} \\
		\end{tabular}}

\end{table*}

\subsection{Ablations}

We perform several ablations to assess the effect of different aspects of our algorithm in our datasets. In Fig.~\ref{fig:ablation}, we show the visual effects of hierarchy consolidation and depth supervision on the result. Clearly, consolidation by removing redundant scaffold Gaussians is essential for visual quality. For urban scenes, we find that the ability for view extrapolation can be diminished if we do not use depth in our datasets. While depth supervision does not improve per-image metrics (see also quantitative single-chunk ablations in Tab.~\ref{tab:ablation}), we note that its inclusion particularly improves the quality of the appearance of roads, which often lack salient features.
We also show the effect of the additional bundle adjustment step for each chunk; the COLMAP hierarchical mapper only provides approximate camera poses that need to be refined. These higher-fidelity poses demonstrably reduce blurriness and increase detail in the resulting novel views.
Finally, we show the effects of exposure compensation and hierarchy optimization: The former removes spurious Gaussians attempting to model differences in intensity between images. The latter improves the quality of intermediate nodes in the hierarchy, leading to sharper shapes and outlines, especially for distant objects.

In addition to the above, for a single chunk of \textsc{Wayve}, we further examined the effect of including or excluding our smooth interpolation during hierarchy optimization. Excluding interpolation both from training and evaluation results in PSNR metrics of 25.21\,db, 24.73\,db and 23.44\,db for granularity settings $\tau_1, \tau_2$ and $\tau_3$, respectively, yielding a clear drop in quality across multiple hierarchy levels (compare with results in Tab.~\ref{tab:comparisons}).

\section{Limitations, Discussion and Future Work}
\label{sec:discuss}

Our results show some visual artifacts. The vast majority of these is due to the input data: bad coverage of the view space, bad calibration, moving distractors (humans, vehicles, especially cars coming to a stop during capture etc). Solving these problems is orthogonal to our method, although radiance fields could help solve some of them. 

For the datasets we show, in most parts of the environment the extrapolation capability of the radiance field is somewhat limited. This is due to limited capture; however it is possible that in future work using good quality priors ~\cite{Nerfbusters2023} could significantly increase the capability for free-viewpoint navigation, even with captures such as the ones we show.

The 3DGS hierarchy could have other applications allowing radiance fields to become a first-class computer graphics representation: it could be used to create scene-graph representations for radiance fields, for animation, collision detection, etc.

Our current solution can significantly reduce the number of Gaussians rendered each frame, thus curbing resource requirements and enhancing flexibility: Instead of a fixed granularity, adding dynamic LOD selection to our prototype would allow to produce optimal quality at a given resource budget. For additional efficiency, a visibility- and distance-based cutoff could be introduced. We leave these considerations to future work.

\section{Conclusion}
\label{sec:concl}

We have presented the first novel view synthesis method that can handle street-level scenes spanning several kilometers in distance, and tens of thousands of input images with real-time rendering.
To allow this, we introduced three contributions: First, an efficient hierarchy of 3D Gaussians, that allows a smooth level-of-detail mechanism for efficient display of massive scenes; Second, the ability to optimize this hierarchy, improving the quality/speed tradeoff and third a divide-and-conquer, chunk-based algorithm for optimizing the hierarchical representation that allows parallel processing of independent chunks, making it possible to fully process a scene of tens of thousands of images in a few hours on a compute cluster. Most importantly, our system allows real-time rendering of such scenes, making the ability to capture and navigate in very large environments accessible to everyone.

\begin{acks}
This research was funded by the ERC Advanced grant FUNGRAPH No 788065 (\MYhref{https://fungraph.inria.fr}{https://fungraph.inria.fr}); B.K. and M.W. acknowledge funding from WWTF (project ICT22-055: Instant Visualization and Interaction for Large Point Clouds). The authors are grateful to Adobe for generous donations, NVIDIA for a hardware donation, the OPAL infrastructure from Universit\'e C\^ote d'Azur and for the HPC resources from GENCI-IDRIS (Grant 2023-AD011014505). Thanks to Fr\'edo Durand and Adrien Bousseau for proof reading and insightful comments, Sebastian Viscay for capturing SmallCity and Nikhil Mohan and colleagues at Wayve for the dataset and overall help.
\end{acks}

\bibliographystyle{ACM-Reference-Format}
\bibliography{references}
\appendix

\begin{appendix}

\vspace{0.2cm}	

We provide several additional details on the implementation of our system. We start with data cleanup, then describe our two-stage pose estimation pipeline for COLMAP, that allows us to calibrate tens of thousands of cameras in several hours wall-clock time. We then describe our regularization strategies for depth and exposure.

\section{Data cleanup}
\label{sec:data}

We capture GoPro time-lapse video on our multi-camera rig (see Fig.~\ref{fig:helmet-bike}). In our long captures there are occasionally blurry frames; we run a sharpness detector (variance of Laplacians) on the images and discard images that are more than 1.5 standard deviations below the mean. This is not foolproof, but helps overall reconstruction quality.

Our scenes are taken in the wild, and contain people and moving vehicles (cars, motorcycles, bicycles). We use Mask R-CNN~\cite{He2017maskrcnn} to identify these classes. We mask out all classes corresponding to people and animals, and detect motion for classes corresponding to vehicles. To identify motion, we check if there exist SfM points with error lower than 1.5 pixels corresponding to the pixels under these masks; for static objects, the density of SfM points is high while for moving objects it is lower. This is quite effective, but not completely accurate, affecting visual quality.

In order to remove vehicle license plates from the captured images we used EgoBlur~\cite{raina2023egoblur}, Mask R-CNN was used to constrain license plate masks to be included in segmented vehicles masks.

\section{Pose estimation} 
\label{sec:pose}

We use the open source COLMAP system~\cite{schoenberger2016sfm} to allow full reproducibility of our method. The standard COLMAP pipeline uses an exhaustive matcher that is prohibitively expensive and often fails on scenes of more than a few thousand images. 
We thus design a custom matcher, similar to COLMAP's sequential matcher but adapted to a multi-camera rig: we match all images from each of the rig's cameras in capture $i$ to all images in capture $i+2^k$ with $k\in\llbracket0,10\rrbracket$. 
We also manually add matches when the capture goes through the same area several times to ensure loop closure. 
Specifically, for a loop closure with indices $i, j$, match all images of all captures with index $i\pm2^l$ to all images of all captures with index $j\pm2^m$ with $(l,m)\in\llbracket0,5\rrbracket^2$. This manual step could be replaced with landmark recognition to identify the loops automatically~\cite{schoenberger2016vote}. 
To further improve the calibration's robustness, we add matches to the 25 nearest neighbour frames given GPS coordinates when available in the image files' EXIF.
GPU-bounded feature extraction and matching take about one and two hours respectively for 40K images.

To estimate camera parameters, poses and SfM points from matches requires running COLMAP's mapper, which is also prohibitively slow for tens of thousands images.
However, COLMAP features a hierarchical mapper that can provide an approximate estimation in reasonable time (for example 110 minutes on two Intel(R) Xeon(R) Gold 6240 CPUs for 40K images).
We set flexible intrinsic model OPENCV and optimize one set of intrinsic per camera then use 
COLMAP's undistort procedure to generate PINHOLE model cameras that are suitable for 3DGS.
As shown in Fig.~\ref{fig:ablation} directly using poses from the hierarchical mapper shows sub-par results.
To improve the local quality of the calibration, we start by running exhaustive feature matching on all cameras selected to optimize the chunk. 
We then retriangulate SfM points and run bundle adjustment using the hierarchical mapper's poses as initialization.

\pagebreak

Finally, we perform a procrustes realignment between the locally bundle-adjusted camera positions and the global cameras positions to compensate for any drift. 
This fine-tune procedure takes between 30 minutes and 5 hours on an NVIDIA V100 16GB depending on the chunk's complexity (number of cameras and SfM points) and greatly improves the quality of the results.

\section{Regularization for Sparse Consumer-level Capture}
\label{sec:reg}

Multi-camera rigs on a vehicle typically provide very sparse coverage of viewing angles, compared to the typical captures recommended for radiance fields; we thus add additional regularization.

\subsection{Depth Regularization}

In contrast to many NeRF datasets, coverage of observed regions is sparse for vehicule based captures we treat in our large datasets.
This is particularly noticeable for the road that is underdetermined: a "mound" in the center of the street explains all the training views perfectly.

We use depth supervision similar to other radiance-fields solutions that have demonstrated its utility~(e.g., ~\cite{Xian2021, Gao-ICCV-DynNeRF} and many others). Instead of using expensive MVS depth, we use recent deep learning based monocular depth estimation and in particular DPT~\cite{Ranftl2021dpt2}. 
Such methods often use a scale and offset invariant loss~\cite{Gao-ICCV-DynNeRF}; in contrast, we use per-frame SfM points' inverse depth given by COLMAP~\cite{schoenberger2016sfm} $\mathbf{D}_\text{SFM}$ to scale DPT inverse depth $\mathbf{D}$. 
We first obtained scale and offset for each set of inverse depth following~\cite{Ranftl2020dpt1}: $t(\mathbf{D}) = \text{median}(\mathbf{D}),\ s(\mathbf{D}) = \frac{1}{M} \displaystyle\sum_{i \in \text{SFM}} |\mathbf{D}_i - t(\mathbf{D})|$ where $\text{SFM}$ are the SfM indices and $M$ is the number of SfM points in the image. 
We then use it to bring the monocular depth to our dataset's scale: \[\mathbf{D}^* = \frac{s(\mathbf{D}_\text{SFM})}{s(\mathbf{D})} \mathbf{D} + t(\mathbf{D}_\text{SFM}) - t(\mathbf{D})  \frac{s(\mathbf{D}_\text{SFM})}{s(\mathbf{D})}\]
To render depth, we swap each Gaussian's colour by the depth $d_i$ of its mean position: $\hat{\mathbf{D}} = \displaystyle\sum_{i=1}^N T_i\alpha_i d_i$.  We regularize by adding a the following loss during training: $\mathcal{L}_\mathbf{D} = \left| {\hat{\mathbf{D}}} - {\mathbf{D}^*}\right|$. 
We also weight the depth loss with exponential decay throughout the optimization starting at $1$ and ending at $0.01$ for per-chunk optimization. 
We propagate depth supervision gradients to each Gaussian's depth and to rendering $\alpha$, which affects each Gaussian's opacity and position in screen space.

\subsection{Exposure optimization} 

In large datasets captured with consumer equipment such as ours, some exposure and small illumination changes are likely to occur.
For example, our GoPro cameras have exposure compensation to allow for change of environment throughout the capture session.
Similar to other radiance fields methods~\cite{martinbrualla2020nerfw,mueller2022instant}, we 
compensate for this by optimizing a per image array. For simplicity, we chose a $3\times4$ affine transformation $E$. We apply the affine transformation to the rendered colour $C$: $C_c = E[C|1]^\intercal$ where $C_c$ is the colour after compensation for exposure change for a given camera. 
We initialize $E$ as identity, and then optimize per-camera exposures with Adam.
The optimization is first performed in the per-chunk step: the learning rate is scheduled with warm-up and exponential decay with initial learning rate (LR) $1e-3$, final LR $1e-4$, a delay multiplier $1e-3$ and $5000$ delay steps. The delay steps ensure that the model is coherent before optimizing exposure. 
For hierarchy post-optimization, we reuse the exposure trained in the initial per-chunk optimization and fix them.

\end{appendix}

\end{document}